\documentclass{bmvc2k}

\usepackage{graphicx}  
\usepackage[percent]{overpic}  
\usepackage{multirow}  
\usepackage{booktabs}
\usepackage{rotating}
\usepackage{amsmath,amssymb}
\usepackage[none]{hyphenat}
\usepackage{geometry}
\usepackage{caption}  
\usepackage{longtable}
\usepackage{wrapfig}
\usepackage{microtype}
\usepackage{xcolor}
\usepackage{hyperref}

\title{RASALoRE: Region Aware Spatial Attention with Location-based Random Embeddings for Weakly Supervised Anomaly Detection in Brain MRI Scans}

\addauthor{Bheeshm Sharma}{bheeshmsharma@iitb.ac.in}{1}
\addauthor{Karthikeyan Jaganathan}{karthikeyanj@iitb.ac.in}{2}
\addauthor{Balamurugan Palaniappan}{balamurugan.palaniappan@iitb.ac.in}{1}

\addinstitution{
	Department of Industrial Engineering \& Operations Research,\\
	IIT Bombay, India.
}
\addinstitution{
	Department of Energy Science and Engineering,\\
	IIT Bombay, India.
}

\runninghead{B. Sharma, K. Jaganathan, B. Palaniappan}{RASALoRE}

\begin{document}
	\sloppy
	
	\maketitle

	\begin{abstract}
		Weakly Supervised Anomaly detection (WSAD) in brain MRI scans is an important challenge useful to obtain quick and accurate detection of brain anomalies, when precise pixel-level anomaly annotations are unavailable and only weak labels (e.g., slice-level) are available. In this work, we propose RASALoRE: Region Aware Spatial Attention with Location-based Random Embeddings, a novel two-stage WSAD framework. In the first stage, we introduce a Discriminative Dual Prompt Tuning (DDPT) mechanism that generates high-quality pseudo weak masks based on slice-level labels, serving as coarse localization cues. In the second stage, we propose a segmentation network with a region-aware spatial attention mechanism that relies on fixed location-based random embeddings. This design enables the model to effectively focus on anomalous regions. Our approach achieves state-of-the-art anomaly detection performance, significantly outperforming existing WSAD methods while utilizing less than 8 million parameters. Extensive evaluations on the BraTS20, BraTS21, BraTS23, and MSD datasets demonstrate a substantial performance improvement coupled with a significant reduction in computational complexity.
		Code is available at  \href{https://github.com/BheeshmSharma/RASALoRE-BMVC-2025/}{\texttt{https://github.com/BheeshmSharma/RASALoRE-BMVC-2025}}.
	\end{abstract}
	
	\section{Introduction}
	\label{sec:RASALoRE_intro}
	
	Anomaly detection in brain MRI scans is a widely recognized task, helpful in timely identification and treatment of related illnesses, but becomes challenging due to limited availability of labeled data with accurate pixel-wise annotations. When slice-level labels are available, weakly supervised anomaly detection (WSAD) methods have become popular alternatives to achieve refined localization. Techniques that make use of Class Activation Map (CAM) \cite{Zhou_2016_CVPR}, including AME-CAM \cite{chen2023ame} and CAE \cite{xie2024weakly}, have shown promise in identifying anomalies in brain MRI scans by utilizing slice-level labels. Similarly, AnoFPDM \cite{che2025anofpdm} has advanced WSAD by leveraging diffusion models.	Despite the strengths of WSAD methods, they struggle with the intricate complexity of brain anatomy, resulting in suboptimal performance when compared to fully supervised methods.
	
	In this work, we propose \textbf{RASALoRE}, an improved weakly supervised anomaly detection framework for brain MRI scans. Operating with only slice-level labels, RASALoRE operates in two phases: a Discriminative Dual-Prompt Training (DDPT) phase which uses pretrained vision-language models for the slice label classification task to generate pseudo weak masks for potential anomalies,  followed by a segmentation model training, leveraging region aware spatial attention mechanism, guided by location-based random embeddings (LoRE). DDPT leverages efficient fine-tuning of visual and language prompts in a vision-language model \cite{radford2021learning} to classify slices as healthy or unhealthy (anomalous) while producing weak supervision to guide RASALoRE’s training. Furthermore, we extend RASALoRE to support multimodality inputs, enhancing its versatility. Extensive experiments on BraTS-type datasets demonstrate that RASALoRE achieves superior performance when compared to state-of-the-art WSAD methods for brain MRI scans.
	
	\section{Proposed Methodology of RASALoRE}
	\label{sec:RASALoRE_methodology}
	In this section, we provide comprehensive details on the two-stage framework adopted for RASALoRE. The first stage, Discriminative Dual Prompt Tuning (DDPT), generates pseudo anomaly masks. In the second stage, we introduce a segmentation network guided by fixed location-based random embeddings (LoRE), enabling precise anomaly localization. In this work, we consider a brain MRI scan image $X \in {\mathbb{R}}^{h\times w}$ obtained as a 2D slice from a 3D brain MRI volume $V\in{\mathbb{R}}^{h\times w\times d}$, where $h, w$ denote the height, width of a slice and $d$ denotes the depth of the volume $V$. Note that though pixel level annotations of anomalies in the slice $X$ are not available, we assume availability of slice-level labels, indicating if a slice contains anomaly or not. 
	
	\subsection{Discriminative Dual Prompt Tuning (DDPT)}
\begin{figure}[htbp]
		\centering
		\includegraphics[width=0.9\linewidth]{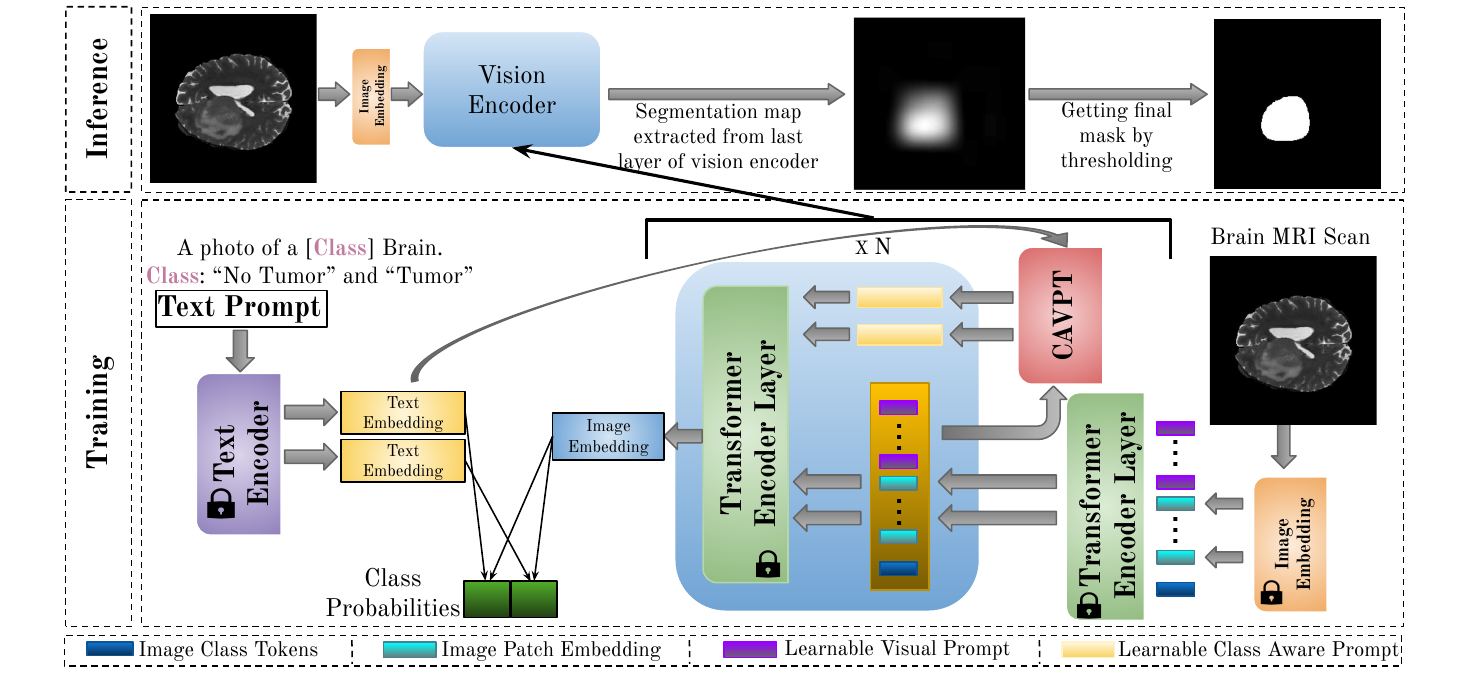}
		\vspace{0.2cm}
		\caption{Overview of Discriminative Dual Prompt Tuning (DDPT)}
		\label{fig:DDPT}
	\end{figure}

	DDPT employs a classification-driven approach to generate coarse anomaly segmentation maps using only weak (slice-level) supervision. 
	By training a discriminative network (e.g. Vision Transformer (ViT) \cite{dosovitskiy2020image} in our case) to classify whether a brain MRI scan image is anomalous or not, we aim to obtain attention maps from the discriminative network, which might contain potential region localization information, guiding the classification task. 
	Then by extracting the relevant attention maps from a suitable layer of the discriminative network, we perform pixel-level anomaly identification. DDPT architecture is illustrated in Figure \ref{fig:DDPT}.
	
	By formulating the anomaly detection task as a binary classification problem, distinguishing between healthy and unhealthy MRI slices, our proposed DDPT builds upon 
	existing works such as CoOP \cite{zhou2022learning}, VPT \cite{jia2022visual}, and DPT \cite{xing2023dual} to leverage learnable vision and text prompts enabling cross-modal interaction.
	We first train learnable text prompts using a frozen text encoder, following CoOP. The prompt is structured as:
	$t = [V]_1 \dots [V]_{\mathtt{M}/2}\,[\text{CLASS}]\,[V]_{\mathtt{M}/2+1} \dots [V]_\mathtt{M}$
	where $[V]_i$ are learnable tokens, $[\text{CLASS}]$ is the class label (\texttt{healthy} or \texttt{unhealthy} in our case), and $\mathtt{M}$ is the prompt length. The corresponding embeddings guide the ViT-based image encoder, which receives patches of input image $X$ along with learnable visual prompts (inspired by VPT). All ViT weights remain frozen; only prompts are trained. We further used CAVPT \cite{xing2023dual}, where visual prompts and text embeddings interact via multi-head attention across ViT layers. This enables context-aware, class-specific attention refinement. A classifier embedded within CAVPT predicts image classes, further guiding embedding refinement. The final layer's refined embeddings are used for classification and segmentation. Class probabilities are computed using cosine similarity between image and text embeddings as:
	\( p_i = \frac{\exp(\cos(q(t_i), f) / \tau)}{\sum_{j=1}^{C} \exp(\cos(q(t_j), f) / \tau)} \), 
	where $q(t_i)$ is the text embedding of prompt $t_i$ for class $i$, $f$ is the image embedding, $C = 2$, and $\tau$ is a temperature parameter.
	
	DDPT minimizes the overall loss given by:
	$\mathcal{L}_{\text{total}} = \eta \mathcal{L}_{\text{ce}}^{\text{ca}} + \mathcal{L}_{\text{ce}}$, 
	where $\mathcal{L}_{\text{ce}}$ is standard cross-entropy between predicted and true labels, 
		$\mathcal{L}_{\text{ce}}^{\text{ca}}$ is an auxiliary cross-entropy loss applied to the output of the class-aware visual prompt generator in CAVPT, using only the query corresponding to the ground-truth class \cite{xing2023dual}. The coefficient $\eta$ balances both terms.
	
	During the inference stage, images are input into the image encoder, while the corresponding textual prompts indicating the presence/absence of anomaly are processed through the text encoder. As the input image propagates through the vision encoder, as shown in the inference part of Figure \ref{fig:DDPT}, attention maps are extracted from the final layer embeddings of DDPT model using thresholding.
	
	\subsection{Region Aware Spatial Attention with Location-based Random Embeddings (RASALoRE)}
	We now describe the training process of our segmentation network RASALoRE, which is guided by fixed location-based random embeddings. This network is trained using the pseudo weak masks generated by the DDPT.

	\begin{figure}[htbp]
		\centering
		\begin{overpic}[width=0.9\textwidth]{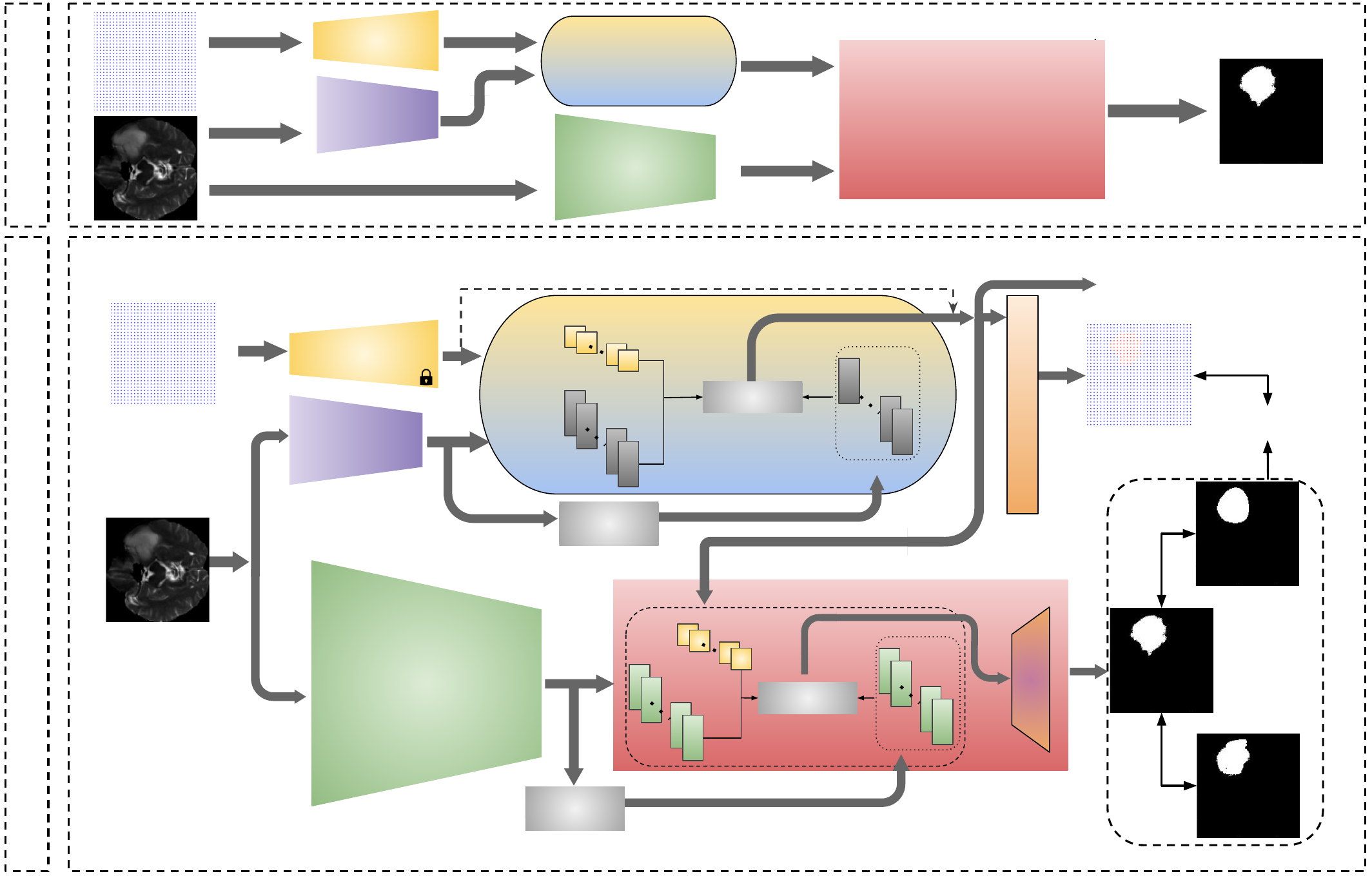}
			\put(24.5, 61.2){\tiny \scalebox{0.8}{\textbf{Positional}}} 
			\put(24, 60.2){\tiny \scalebox{0.8}{\textbf{Embeddings}}}
			\put(24, 55){\scalebox{0.6}{\textbf{Refiner}}}
			\put(43, 58.8){\scalebox{0.7}{\textbf{RASA}}}
			\put(43.5, 52){\scalebox{0.7}{\textbf{Image}}} 
			\put(42.5, 50){\scalebox{0.7}{\textbf{Encoder}}} 
			\put(68, 55.5){\scalebox{0.7}{\textbf{Mask}}}
			\put(67, 53.5){\scalebox{0.7}{\textbf{Decoder}}}
			\put(86, 50){\scalebox{0.6}{\textbf{Predicted Mask}}}
			\put(1,51){\rotatebox{90}{\scalebox{0.7}{\textbf{Inference}}}}
			\put(1,20){\rotatebox{90}{\scalebox{0.7}{\textbf{Training}}}}
			\put(6.5,33){\tiny \scalebox{0.9}{\textbf{Candidate Prompt}}}
			\put(10,31.7){\tiny \scalebox{0.9}{\textbf{Points}}}
			\put(6, 17){\tiny \textcolor{black}{\textbf{Brain MRI Scan}}}
			\put(6, 15.7){\tiny \textcolor{black}{\textbf{(With Anomaly)}}}	
			\put(23, 41){\small \scalebox{0.9}{$E_{\text{cpp}}$}}
			\put(22.3, 38.3){\tiny \textbf{Positional}}
			\put(22, 37){\tiny \textbf{Embeddings}}
			\put(21.2, 31){\scalebox{0.7}{\textbf{Refiner} $R_{\rho}$}}
			\put(24, 14.7){\small \textbf{Image}} 
			\put(24, 12.2){\small \textbf{Encoder}} 
			\put(34, 13.7){\small $U_\varrho$} 
			\put(36.5, 37.5){\tiny \textbf{Query}} 
			\put(37, 31){\tiny \textbf{Key}} 
			\put(62, 29.2){\tiny \textbf{Value}} 
			\put(51.7, 34.5){\tiny \textbf{Attention}} 
			\put(46.5, 20){\tiny \textbf{Query}} 
			\put(46, 9){\tiny \textbf{Key}} 
			\put(64.2, 9.5){\tiny \textbf{Value}} 
			\put(55.7, 12.5){\tiny \textbf{Attention}} 
			\put(73.1, 25){\tiny \textbf{FFN}} 
			\put(41.5, 26.3){\scalebox{0.7}{\tiny \textbf{Added with}}}
			\put(42, 25.3){\scalebox{0.7}{\tiny \textbf{Gaussian}}}
			\put(43, 24.3){\scalebox{0.7}{\tiny \textbf{Noise}}}
			\put(39, 5.5){\scalebox{0.7}{\tiny \textbf{Added with}}}
			\put(39.5, 4.5){\scalebox{0.7}{\tiny \textbf{Gaussian}}}
			\put(40.5, 3.5){\scalebox{0.7}{\tiny \textbf{Noise}}}
			\put(74.9, 10.4){\rotatebox{90}{\tiny \textbf{Upsampler}}} 
			\put(46, 40){\small \textbf{RASA}} 
			\put(52, 22.8){\scalebox{0.4}{\textbf{Enriched Spatial Point Embedding $\xi_{ESPE}$}}} 
			\put(53.5, 19.6){\small \textbf{Mask Decoder}} 
			\put(45, 43){\tiny \textbf{Residual Connection}} 
			\put(88.5, 15){\tiny $M_{\text{ANO}}$} 
			\put(82, 27.5){\tiny $M_{\text{DDPT}}$} 
			\put(82.8, 4){\tiny $M_{\text{SAM}}$} 
			\put(77, 4.5){\scalebox{0.6}{\tiny $ \boxed{\textcolor{red}{\mathcal{L}_{\text{Dec}}}}$}} 
			\put(80, 43){\scalebox{0.6}{\tiny \boxed{\textcolor{red}{\mathcal{L}_{\text{Struct}}}}}} 
			\put(91, 37.8){\scalebox{0.6}{\tiny $ \boxed{\textcolor{red}{\mathcal{L}_{\text{PA}}}}$}} 
			\put(75.8, 35){\scalebox{0.32}{\textbf{Reshape}}} 
			\put(88, 33){\scalebox{0.5}{\textbf{Point Activation}}} 
			\put(89.8, 31.8){\scalebox{0.5}{\textbf{Mask} (\scalebox{1.0}{$M_{\text{PA}}$})}} 
			\put(81, 40.8){\tiny $M_{\text{FFN}}$} 
			
			\put(81.5, 12.2){\tiny \textcolor{yellow}{\textbf{Predicted}}}
			\put(88, 3){\tiny \textcolor{yellow}{\textbf{MedSAM}}}
			\put(86, 1){\scalebox{0.8}{\tiny \textbf{(Obtained using}}}
			\put(94.5, 1){\scalebox{0.8}{\tiny $M_{\text{DDPT}}$\textbf{)}}}
			\put(89, 21.5){\tiny \textcolor{yellow}{\textbf{DDPT}}}
		\end{overpic}
		\vspace{0.25cm}
		\caption{Overview of RASALoRE Architecture}
		\label{fig:RASALoRE_arch}
	\end{figure}
\begin{figure}[h!]
		\centering
		\begin{overpic}[width=0.9\textwidth]{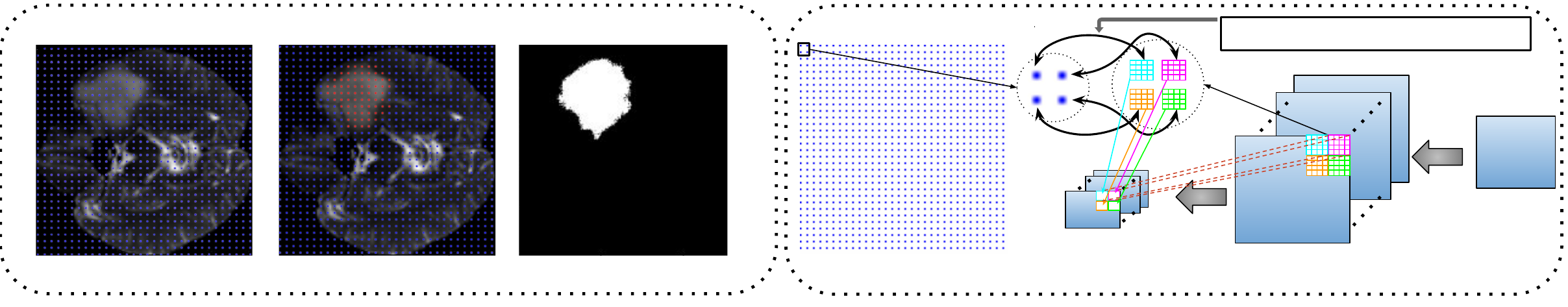}
			\put(25,-2){\large\textbf{(a)}}
			\put(75,-2){\large\textbf{(b)}}
			\put(52, 2.5){\scalebox{0.5}{\textbf{Candidate Prompt}}} 
			\put(53, 1.3){\scalebox{0.5}{\textbf{Point Locations}}} 
			\put(66, 3.8){\scalebox{0.5}{\textcolor{black}{\textbf{Refiner Pooled}}}}	
			\put(68, 2.6){\scalebox{0.5}{\textcolor{black}{\textbf{Output}}}}	
			\put(78, 2.8){\scalebox{0.5}{\textcolor{black}{\textbf{Refiner Intermediate}}}}
			\put(79, 1.6){\scalebox{0.5}{\textcolor{black}{\textbf{Representations}}}}
			\put(94.5, 6.3){\scalebox{0.5}{\textcolor{black}{\textbf{Refiner}}}}
			\put(95, 5.1){\scalebox{0.5}{\textcolor{black}{\textbf{Input}}}}
			\put(78, 17.57){\scalebox{0.37}{\textcolor{black}{\textbf{Interaction between Candidate prompts}}}}  
			\put(78.5, 16.57){\scalebox{0.37}{\textcolor{black}{\textbf{and Region Representations (In RASA)}}}}
		\end{overpic}
		\vspace{0.5cm}
		\caption{(a) Left: Candidate prompt point locations (in blue) overlaid as grid on input image, center: point activation mask (red denoting active and blue denoting inactive points) overlaid on input image, right: weak anomaly mask corresponding to input image. (b) Refiner Module.}
		\label{fig:LPA_Refiner}
	\end{figure}

	\textbf{LoRE:} Our approach centers on using location-based random embeddings (LoRE), where specific spatial positions on the input image $X$ are designated as candidates. For $X$, we first generate a $\sqrt{k} \times \sqrt{k}$ grid of $k$ evenly spaced point coordinates across both rows and columns (see Figure \ref{fig:LPA_Refiner} (a)). 
	Each grid point forms a candidate prompt point (CPP) in our approach. Our central idea is to enrich these grid points with representational information from the brain MRI images so that a select few of these grid points will serve as potential prompts for a particular image, eliciting the corresponding anomaly-related information. Once the CPP's $(x,y)$ coordinates are fixed, they are normalized to the range \([-1, 1]\), and \({d}\)-dimensional location embeddings are derived for each point's coordinates based on sinusoidal transformations. Unlike existing prompt encoders (e.g. MedSAM \cite{Ma_2024}) where the location embeddings are learnable, our LoRE provide fixed, non-learnable encodings that are independent of dataset-specific biases. The CPPs and their LoRE denoted by \( E_{\text{cpp}} \in {\mathbb{R}^{k \times d}} \), remain fixed throughout the training as well as testing process, and are shared by all train/test set images. Since our methodology heavily relies on accurate and fixed CPP locations, ensuring effective regional information sharing is crucial. Corresponding to the CPPs,  image representations are obtained from a refiner (denoted by $R_\rho$, see figure \ref{fig:RASALoRE_arch}). The refiner processes the input $X$ using a series of convolutions and outputs  $R_\rho(X) \in {\mathbb{R}^{\sqrt{k} \times \sqrt{k} \times d}}$, 
	containing $k$ pixels corresponding to the number of CPP locations. Each pixel in the output $R_\rho(X)$ of the refiner corresponds to a particular region in the image representation, enabling each CPP to effectively share information with, and learn from the characteristics of its corresponding neighborhood region in the image, enhancing the model's ability to capture spatial dependencies. The refiner module is illustrated in Figure \ref{fig:LPA_Refiner} (b). 
	
	\textbf{RASA:}  The location-based random embeddings $E_{\text{cpp}}$ interact with spatial information of $X$ obtained as $R_\rho(X)$ from the Refiner, in a module called Region Aware Spatial Attention (RASA) module, to result in enriched spatial point embeddings \( \xi_{\text{ESPE}} \in \mathbb{R}^{k \times d} \), corresponding to the $k$ CPPs.  RASA primarily comprises a multi-head attention (MHA) \cite{vaswani2017attention} computation denoted by $\text{MHA}_{\text{RASA}}(Q, K, V)$. The CPPs' positional embeddings $E_{\text{cpp}}$ form the query $Q$ for $\text{MHA}_{\text{RASA}}$. Intermediate representations $R_\rho(X)$ from Refiner act as key $K$ in RASA. The value $V$ of $\text{MHA}_{\text{RASA}}$ is based on the perturbed representation $R_\rho(X)+\epsilon$, where $\epsilon \sim \mathcal{N}(\textbf{0},\textbf{I})$ represents Gaussian noise. This noise addition is being performed to improve the robustness of the attention module. Further a residual path adds information of $E_{\text{cpp}}$ to the output of  $\text{MHA}_{\text{RASA}}$.

\textbf{Mask Decoder:} The enriched spatial point embeddings $\xi_{\text{ESPE}}$ are then fed into a mask decoder, a feed forward network, and a structural loss computation module. 
The mask decoder performs MHA (denoted by  $\text{MHA}_{\text{Dec}}(Q,K,V)$), allowing interactions between query $Q=\xi_{\text{ESPE}}$ denoting the region aware enriched spatial point embeddings from RASA and key $K=U_\varrho(X)$ obtained as the image representations from an image encoder. In our model, the image encoder $U_\varrho$ provides an intermediate feature representation of the input image $X$, and it contains four encoder blocks, whose design is based on that of the encoder of UNet \cite{DBLP:journals/corr/RonnebergerFB15}.
The output from  $\text{MHA}_\text{Dec}$, after a suitable upsampling step then provide the anomaly mask predictions $M_{ANO} \in {\mathbb{R}^{h \times w}}$. 
	
	The weak segmentation mask \( M_{DDPT} \) produced by DDPT provides an approximate delineation of the anomalous regions. We observed that the weak mask $M_{DDPT}$ has a smooth boundary; nevertheless, it provides a better localization of the interior of the potential anomalous regions. We further use the weak mask from DDPT to prompt a pre-trained MedSAM \cite{Ma_2024} model and use the resultant weak mask $M_{SAM}$ as another weak supervision signal. Although the masks obtained from MedSAM are also weak, they capture boundary-level information of the potential anomalous regions to some extent. We design a custom loss function to compare the output mask from the mask decoder with the pseudo weak masks obtained from DDPT and DDPT-prompted MedSAM. 
	Our loss function is of the form:
	\begin{align}
		\mathcal{L}_{\text{Dec}} &= \text{ELDice}\left( M_{ANO}, G_\sigma(M_{DDPT}) \right) 
		+ \gamma \cdot \text{ELDice} \left( M_{ANO}, G_\sigma^{-1}(M_{SAM}) \right) \nonumber \\
		&\quad + \frac{\alpha}{p} \cdot \left( M_{ANO} \odot (1 - M_{DDPT}) \right) \nonumber \\
		&\quad + \beta \cdot \text{ELDice} \left( (1 - M_{ANO}) \odot (1 - M_{DDPT}), 
		G_\sigma(1 - M_{DDPT}) \right),
		\label{eq:decoder_loss_new}
	\end{align}	
where the ELDice (Exponential-Logarithmic-Dice) loss \cite{Wong_2018} between a predicted mask \( P \) and a binary ground truth mask \( GT \) is: \(\text{ELDice}(P, GT) = (-\ln((2\mathtt{I} + \epsilon)/(\mathtt{U} + \epsilon)))^\phi\), where \(\mathtt{I}=|P \cap GT|\) denotes the number of pixels common to \(P\) and \(GT\) and \(\mathtt{U}=|P| + |GT|\) is the total number of pixels in \(P\) and \(GT\), \( \epsilon > 0 \) is a small smoothing constant and \( \phi = 0.3 \).     

	In eq. \eqref{eq:decoder_loss_new}, the first two loss terms indicate comparison of predicted mask $M_{ANO}$ from mask decoder with the weak supervision masks $M_{DDPT}$ and $M_{SAM}$, using ELDice loss. 
	For the weak mask $M_{DDPT}$, a Gaussian filter $G_\sigma$ provides larger weights towards the center of the predicted anomalous region in mask and the weights gradually decrease towards the boundary. Conversely for weak mask $M_{SAM}$, an inverse Gaussian filter $G_\sigma^{-1}$ assigns lower weights to the center, and progressively increasing weights toward the boundaries. These filters are used to encourage the model to focus on the boundary regions, where structural details are more prominent, helping it learn fine-grained shape and edge information that may be overlooked when only center-weighted supervision (as in DDPT) is used.
	
	The second loss term is weighed using a particular factor $\gamma = Dice(M_{SAM}, M_{DDPT})$, which allows $M_{SAM}$ information to contribute to the loss only when $M_{SAM}$ and $M_{DDPT}$ masks overlap well. To control False Positives (FPs) in $M_{ANO}$, we introduce the last two terms in $\mathcal{L}_{\text{Dec}}$. 
	{The third loss term in eq. \eqref{eq:decoder_loss_new} denotes mean confidence of false positive pixels in $M_{ANO}$, where $p$ denotes number of pixels in $M_{ANO}$. The fourth loss term in eq. \eqref{eq:decoder_loss_new} calculates the ELDice score between the true negatives of prediction $M_{ANO}$ and the DDPT-based mask $M_{DDPT}$, aiming to reduce false positives by improving the true negative performance. The notation $\odot$ in eq. \eqref{eq:decoder_loss_new} denotes elementwise multiplication.} 
	
	\textbf{FFN Details:} The $\xi_{\text{ESPE}}$ output from RASA module is also fed to a simple feed-forward network, which projects $\xi_{\text{ESPE}}$ to a grid structured anomaly mask $M_{FFN} \in {\mathbb{R}^{\sqrt{k} \times \sqrt{k}}}$. The mask $M_{FFN}$ contains activations corresponding to the CPPs' locations, indicating whether these candidate prompt points potentially correspond to an anomaly or not. To compare $M_{FFN}$, we extract mask information from the weak DDPT mask, corresponding to the grid-point structure of CPPs, resulting in a point activation mask $M_{PA}$ (see Figure \ref{fig:LPA_Refiner} (a)), and construct the following loss function:
	$\mathcal{L}_{\text{PA}} = \text{ELDice} \left( M_{\text{FFN}}, M_{PA} \right)$, 
	where ELDice loss is used to compare $M_{FFN}$ and corresponding point activations based weak mask $M_{PA}$ derived from $M_{DDPT}$. 
	
	\textbf{Structural loss for embeddings:} The $\xi_{\text{ESPE}}$ from the RASA module is also fed into a structural loss computation module, which aims to attain similarity among the embeddings corresponding to CPPs representing anomalies. This structural loss is given as: 
	$\mathcal{L}_{\text{Struct}} = \text{MSE} \left( \xi_{\text{ESPE}}^{A}, \textbf{1} \right) +  \text{MSE} \left( \xi_{\text{ESPE}}^{IA}, \textbf{-1} \right)$
	Here $\xi_{\text{ESPE}}^{A}$ denoting enriched spatial point embeddings corresponding to active points in the point activation mask \( M_{\text{PA}} \) (where \( M_{\text{PA}} = 1 \)) are forced towards value $\textbf{1}$, and $\xi_{\text{ESPE}}^{IA}$ denoting embeddings corresponding to inactive points (where \( M_{\text{PA}} = 0 \)) are forced towards $\textbf{-1}$.
	$\mathcal{L}_{\text{Struct}}$ aims to obtain a distinction between the components of embeddings corresponding to active and inactive points in $M_{PA}$, helping the model learn better enriched spatial point embeddings.
	
	The overall network of RASALoRE (shown in Figure \ref{fig:RASALoRE_arch}) is trained by minimizing $\mathcal{L} = \mathcal{L}_{\text{Dec}} + \mathcal{L}_{\text{PA}} + \mathcal{L}_{\text{Struct}}$. 
	During inference on an arbitrary test image $\hat{X}$, image embeddings $U_\varrho(\hat{X})$ obtained from the image encoder and enriched LoRE obtained from the RASA module, when passed to mask decoder, provide the desired anomaly segmentation mask prediction. 
	
	\subsection{Multimodality RASALoRE}
		Further we extended RASALoRE to support multiple MRI modalities. Assuming that RASALoRE was pretrained on modality $m \in \mathcal{M}$ ($\mathcal{M}$ being the set of available MRI modalities), 
		we designate $m$ as a bridge modality. Using the pretrained model, we extract enriched embeddings, denoted as $\xi_{\text{ESPE}}^{\text{bridge}}$, from all train data slices via the RASA module. These embeddings serve as reference targets to align embeddings from other modalities into a shared feature space, ensuring consistent and robust representation across modalities.
		
		To facilitate multimodal integration, we augment our architecture with \( | \mathcal{M} |\) distinct sets of CPPs and their corresponding LoRE, 
		and associate each with its own dedicated RASA module. Importantly, the encoder, refiner, and mask decoder components remain shared across modalities, enabling parameter-efficient multimodal learning. 
		Moreover, at inference time, predictions can be obtained using any individual modality or combinations thereof without requiring all modalities simultaneously. Crucially, the total number of parameters engaged during inference remains similar across different modalities, as only the relevant RASA module is activated based on the available modality. 
		
		To ensure cross-modality alignment, we introduce an additional loss that encourages enriched embeddings from all modalities to align closely with the reference bridge embedding $\xi_{\text{ESPE}}^{\text{bridge}}$. Let $\xi_{\text{ESPE}}^{(j)}$ represent the enriched embedding for modality $j \in \mathcal{M}$.
		We define the bridge alignment loss as \( \mathcal{L}_{\text{align}} = \sum_{j \in \mathcal{M}} \left\| \xi_{\text{ESPE}}^{(j)} - \xi_{\text{ESPE}}^{\text{bridge}} \right\|_2^2 \), 
		which promotes a unified feature space across modalities. The overall network of Multi-modality RASALoRE is trained by minimizing $\mathcal{L} = \mathcal{L}_{\text{Dec}} + \mathcal{L}_{\text{PA}} + \mathcal{L}_{\text{Struct}} + \lambda_{\text{align}} \cdot \mathcal{L}_{\text{align}}$. where $\lambda_{\text{align}}$ controls the contribution of the alignment objective.
		
		\section{Related Works}
		\label{sec:RASALoRE_relatedwork}
		
		\textbf{Weakly Supervised Approaches:}
		Existing weakly supervised approaches, such as CAM-based methods \cite{Zhou_2016_CVPR}, have been extensively studied and extended to improve localization under limited supervision. CAE \cite{xie2024weakly} employs topological data analysis to extracted class-related features, thereby enhancing focus on anomalous regions. AME-CAM \cite{chen2023ame} introduces a multi-exit classifier architecture that captures internal activation maps at multiple depths and uses attentive feature aggregation to produce refined attention maps. LA-GAN \cite{tao2023lagan} utilizes a three-stage approach, comprising classifier training, pseudo map generation, and GAN-based generative training. A similar GAN-based approach is used in volumetric sense in Yoo et al \cite{Yoo2025GenerativeAI}. 
		Kim et al.~\cite{kim2022bridging} propose aligning image-level features with class-specific weights to recover less discriminative regions, in non-medical imaging data. Similarly, transformer-based methods such as TS-CAM \cite{gao2021ts} and SCM \cite{bai2022weakly} enable patch tokens to become object-category aware, which improves localization performance. Our DDPT method is similar in spirit to existing CAM-based methods; however, by using vision-language prompt tuning, DDPT achieves improved weak annotations. 
				
		\textbf{Reconstruction-based Approaches:}
		Several reconstruction-based methods, also referred to as Unsupervised Anomaly Detection (UAD) techniques, utilize autoencoders \cite{baur2020autoencodersunsupervisedanomalysegmentation, kascenas2022denoising}, variational autoencoders \cite{marimont2020anomalydetectionlatentspace}, and diffusion models (e.g. Denoising Diffusion Probabilistic Models (DDPMs) \cite{ho2020denoising}, Patch-based DDPMs (pDDPMs) \cite{behrendt2024patched}, Masked DDPMs (mDDPMs) \cite{iqbal2023unsupervised}, Conditional DDPMs (cDDPMs) \cite{behrendt2023guided}). For a comprehensive survey on autoencoder and variational autoencoder based methods, see \cite{kascenas2023anomaly, pinon2024unsupervised}. 
		In addition to these methods, several transformer-based models have been explored. \cite{pinaya2021unsupervisedbrainanomalydetection} employs VQVAE combined with transformers, while \cite{rashmi2024anoswinmae} adopts a Swin Transformer-based masked encoder. These methods typically rely on volume-level labels and are trained exclusively on healthy brain MRI scans, with inference performed on unhealthy cases. In contrast, our approach employs frame-level labels, placing it under the weakly supervised learning paradigm. We note that AnoFPDM \cite{che2025anofpdm} also proposes a reconstruction-based approach using a diffusion model with classifier-free guidance; however, it uses frame-level labels, making it a weakly supervised approach. Unlike traditional approaches that rely on iterative reconstruction, AnoFPDM utilizes the forward diffusion process to identify anomalies.

		\textbf{Prompt-driven Approaches:}
		Human-driven, prompt-based methods have also gained popularity in anomaly detection, for querying large foundational models like MedSAM \cite{Ma_2024} and MedSAM2 \cite{zhu2024medicalsam2segment}. These models rely on prompts, which offer stronger supervision compared to frame-level supervision in WSAD. We propose a prompt guidance for MedSAM and MedSAM2 using DDPT-generated pseudo weak masks and performed experiments.

		\section{Experiments and Results}
		\label{sec:RASALoRE_expts}
		\textbf{Experimental Setup:} In this section, we present the experimental setup and results related to DDPT and RASALoRE. 
		We conducted all experiments using PyTorch 2.0.1 framework on a Linux system, with a NVIDIA GeForce A6000 graphics card, having 48GB of memory. 
        For DDPT training, we employed the SGD optimizer with a learning rate of 0.01, weight decay set at $5\times 10^{-4}$, and momentum at $0.9$ for model training as used in \cite{xing2023dual}. We set temperature coefficient $\tau=0.07$ from \cite{wu2018unsupervised} and set $\eta=0.3$ based on ablation. For RASALoRE training, we utilized the SGD optimizer with a learning rate lr = 0.01, a momentum of 0.9 and a batch size of 16. Additionally, we employed a linear learning rate scheduler that halves the lr after every 20 epochs. Since the objective is weakly supervised and lacks pixel-level annotations, we use a standard threshold of 0.5 to obtain the segmentation mask.
		Notably, our model training needs only around 12GB of memory on a single GPU. For our training, we choose the number of CPP locations \(k = 1024\), overlaid as $32 \times 32$ grid over every MRI image of size \(256 \times 256\). The embedding dimension \(d\) is set to 256. The values of \(\alpha\) and \(\beta\) are both set to 0.6, determined based on ablation study. In both RASA and the Mask Decoder, we use 4 heads in the multi-head attention (MHA).

		\textbf{Dataset and Preprocessing:} We conducted our experiments on four datasets: BraTS20 \cite{bakas2017advancing,bakas2018identifying,menze2014multimodal}, BraTS21 \cite{baid2021rsna,bakas2017advancing,menze2014multimodal}, BraTS23 \cite{kazerooni2024braintumorsegmentationbrats}, and MSD \cite{Antonelli_2022} (which is based on BraTS16 and 17 challenges).
		These datasets are provided in volumetric NIFTI \cite{bakas2018identifying} format, and we extracted 2D slices from the T2 modality volumes. BraTS20, BraTS21, BraTS23 and MSD datasets contain 369, 1251, 1251, and 484 samples in total, respectively. All datasets are split patient-wise into training and testing sets, with 80\% of the data allocated for training and the remaining 20\% for testing. 
		
		While extracting frames from volumetric data, several preprocessing steps were applied. The first and last 15 frames were excluded as they typically contain minimal information and do not feature the brain region prominently. Subsequently, the frames were cropped to remove unnecessary black regions. During training, several data augmentations were applied to improve the model's generalization. First, gamma correction was applied by squaring each pixel's value and normalizing it between 0 and 255 on the input image slices. Then, random rotations within a range of \(-90^\circ\) to \(90^\circ\), and additional random horizontal and vertical flips were applied to both the image and the masks. The brightness and contrast of the image were also randomly adjusted within a range of 0.8 to 1.2. To enhance robustness against noise, Gaussian noise $\epsilon \sim \mathcal{N}(\textbf{0}, \textbf{10}I)$ was incorporated into the images.
		
		\begin{table}[h]
			\centering
			\small  
			\resizebox{\textwidth}{!}{  
				\setlength{\tabcolsep}{4pt}  
				\begin{tabular}{l l c c c c c c c c}
					\toprule
					\textbf{Approach} & \textbf{Method} & \multicolumn{2}{c}{\textbf{BraTS20}} & \multicolumn{2}{c}{\textbf{BraTS21}} & \multicolumn{2}{c}{\textbf{BraTS23}} & \multicolumn{2}{c}{\textbf{MSD}} \\
					\cmidrule(lr){3-4} \cmidrule(lr){5-6} \cmidrule(lr){7-8} \cmidrule(lr){9-10}
					& & \textbf{Dice \( \uparrow \)} & \textbf{AUPRC \( \uparrow \)} & \textbf{Dice \( \uparrow \)} & \textbf{AUPRC \( \uparrow \)} & \textbf{Dice \( \uparrow \)} & \textbf{AUPRC \( \uparrow \)} & \textbf{Dice \( \uparrow \)} & \textbf{AUPRC \( \uparrow \)} \\ \midrule
					
					\multirow{3}{*}{UAD} 
					& AE \cite{baur2020autoencodersunsupervisedanomalysegmentation, kascenas2022denoising} & 14.26\%  & 10.23\%  & 11.83\%  & 8.01\%  & 17.09\%  & 7.41\%  & 14.96\%  & 7.07\%  \\
					& DAE \cite{kascenas2022denoising} & 21.33\%  & 18.89\%  & 14.38\%  & 14.59\%  & 34.16\%  & 21.18\%  & 32.13\%  & 20.77\%  \\
					& VQVAE \cite{pinaya2021unsupervisedbrainanomalydetection} & 17.27\%  & 12.04\%  & 25.69\%  & 17.67\%  & 20.67\%  & 38.48\%  & 19.44\%  & 33.78\%  \\
					\cmidrule(lr){1-10}
					\multirow{10}{*}{WSAD} 
					& TS-CAM \cite{gao2021ts} & 6.13\%  & 7.92\%  & 6.74\%  & 8.35\%  & 9.13\%  & 9.36\%  & 7.81\%  & 8.47\%  \\
					& CAE \cite{xie2024weakly} & 26.36\%  & 17.48\%  & 23.82\%  & 14.20\%  & 46.98\%  & 60.11\%  & 27.96\%  & 18.64\%  \\
					& LA-GAN \cite{tao2023lagan} & 34.14\%  & 28.48\%  & 42.75\%  & 38.82\%  & 40.57\%  & 43.65\%  & 33.63\%  & 27.7\%  \\
					& AME-CAM \cite{chen2023ame} & 52.22\%  & 37.39\%  & 50.43\%  & 37.85\%  & 39.19\%  & 26.47\%  & 51.91\%  & 40.34\%  \\
					& AnoFPDM \cite{che2025anofpdm} & 37.18\%  & 38.78\%  & 41.83\%  & 47.89\%  & 49.28\%  & 57.04\%  & 43.42\%  & 50.08\%  \\
					& Yoo et al. (T2)\cite{Yoo2025GenerativeAI} & 22.76\%  & 13.12\%  & 11.94\%  & 8.64\%  & 12.2\%  & 8.9\%  & 12.09\%  & 8.79\%  \\
					& Yoo et al. (All)\cite{Yoo2025GenerativeAI} & 49.91\%  & 38.64\%  & 63.33\%  & 50.28\%  & 23.41\%  & 12.99\%  & 47.81\%  & 35.93\%  \\
					\cmidrule(lr){2-10}
					& DDPT & 61.53\% & 46.89\% & 51.72\% & 35.79\% & 48.59\% & 31.66\% & 48.71\% & 33.87\% \\
					& M2+DDPT(p) \cite{zhu2024medicalsam2segment} & 32.57\% & 24.10\% & 34.73\% & 25.55\% & 48.52\% & 39.93\% & 43.43\% &  34.75\% \\
					&  M2+DDPT(b) \cite{zhu2024medicalsam2segment} & 35.58\% & 25.44\% & 37.24\% & 26.54\% & 53.54\% & 43.15\% & 45.90\% & 35.78\% \\
					& M+DDPT(p) \cite{Ma_2024} & 37.66\% & 26.29\% & 43.44\% & 29.49\% & 39.22\% & 25.39\% & 38.08\% & 25.46\% \\
					&  M+DDPT(b) \cite{Ma_2024} & 43.44\% & 33.36\% & 51.19\% & 36.54\% & 50.40\% & 33.66\% & 46.46\% & 33.02\% \\
                    & RASALoRE & \textbf{70.57\%} & \textbf{74.74\%} & \textbf{70.85\%} & \textbf{75.05\%} & \underline{70.79\%} & \underline{71.18\%} & \textbf{61.37\%} & \underline{63.71\%} \\
                    & R.Without MedSAM & \underline{69.8\%} & \underline{73.06\%} & \underline{68.87\%} & \underline{74.26\%} & \textbf{74.22\%} & \textbf{80.70\%} & \underline{61.34\%} & \textbf{67.08\%} \\
					
					\bottomrule
				\end{tabular}
			}
			\vspace{0.5em}  
                \caption{Comparison of quantitative results. Abbreviations: M+DDPT(b) = MedSAM+DDPT (box), M+DDPT(p) = MedSAM+DDPT (point), M2+DDPT(b) = MedSAM2+DDPT (box), M2+DDPT(p) = MedSAM2+DDPT (point), and R.Without MedSAM = RASALoRE Without MedSAM. The best values of each metric are in bold, and second best values are underlined.}
			\label{tab:Quantitative_same_domain_results}
		\end{table}
		
		\textbf{Empirical Results:}
		Table~\ref{tab:Quantitative_same_domain_results} provides a comparative evaluation of our proposed RASALoRE against other CAM-based WSAD methods, including CAE \cite{xie2024weakly}, AME-CAM \cite{chen2023ame}, TS-CAM \cite{gao2021ts}, LA-GAN \cite{tao2023lagan}, and approaches in Yoo et al. \cite{Yoo2025GenerativeAI} (using both the T2 modality and the combined modalities). Further we compared with reconstruction based models such as Autoencoders (AE) \cite{baur2020autoencodersunsupervisedanomalysegmentation}, Denoising Autoencoders (Denoising-AE) \cite{kascenas2022denoising}, Vector Quantized Variational Autoencoders (VQVAE) \cite{marimont2020anomalydetectionlatentspace} and AnoFPDM \cite{che2025anofpdm}. 
		All competing methods were reproduced with same baseline settings to ensure a fair comparison.

        We observe that classical reconstruction-based models such as AE, DAE, and VQVAE achieve relatively low Dice and AUPRC values across all benchmarks, reflecting their limited ability to capture complex tumor appearances. Recent CAM-based methods such as CAE, LA-GAN, AME-CAM, and AnoFPDM show improvements, yet their performance fluctuates considerably between datasets, with notable drops in either Dice or AUPRC. Yoo et al. Approaches in \cite{Yoo2025GenerativeAI}, which operate directly on 3D volumetric data using a three-stage training pipeline and pseudo maps to guide the final segmentation network, demonstrates decent performance on BraTS20, BraTS21, and MSD. However, its generalizability remains limited, as evident from the notable performance drop on BraTS23. In contrast, RASALoRE demonstrates strong and stable performance across all datasets (BraTS20, BraTS21, BraTS23, and MSD), achieving significant improvements in critical metrics like Dice Score and AUPRC, particularly important for segmentation tasks. 
        
        We further analyze the performance of MedSAM-integrated variants \cite{Ma_2024, zhu2024medicalsam2segment}(M+DDPT and M2+DDPT), prompted using point or box, derived from DDPT’s weak masks. While these combinations improve basic reconstruction or CAM-based methods, their performance remains significantly below RASALoRE, suggesting that applying powerful foundation models like MedSAM in a plug-and-play manner is insufficient.
        Further, the variant of RASALoRE (R.Without MedSAM), relying solely on DDPT-generated weak masks, achieves results that are often second only to those of full RASALoRE model. This observation validates the reliability of DDPT’s weak supervision and indicates that RASALoRE does not critically depend on MedSAM-based masks.		
		
		\textbf{Qualitative results:} 
        Figure \ref{fig:RASALoRE_Qualitative} presents the visualization of anomaly masks predicted by our model along with those generated by comparative methods. 
        Reconstruction-based models (AE, DAE, VQVAE) fail to capture the irregular tumor boundaries, often producing blurred or incomplete segmentations. CAM-based methods (CAE, LA-GAN, AME-CAM), approaches in Yoo et al., and AnoFPDM, exhibit partial improvements but tend to miss finer structural details or introduce false positives. DDPT-guided MedSAM and MedSAM2 improve localization by leveraging prompt-based supervision from DDPT. In contrast, RASALoRE (with and without MedSAM) produces sharper and more accurate anomaly delineations,  
        illustrating the robustness of RASALoRE in handling diverse tumor morphologies and its ability to generalize better than prior reconstruction/CAM-based approaches.

		\begin{figure}[!t]
			\centering
			\includegraphics[width=\linewidth]{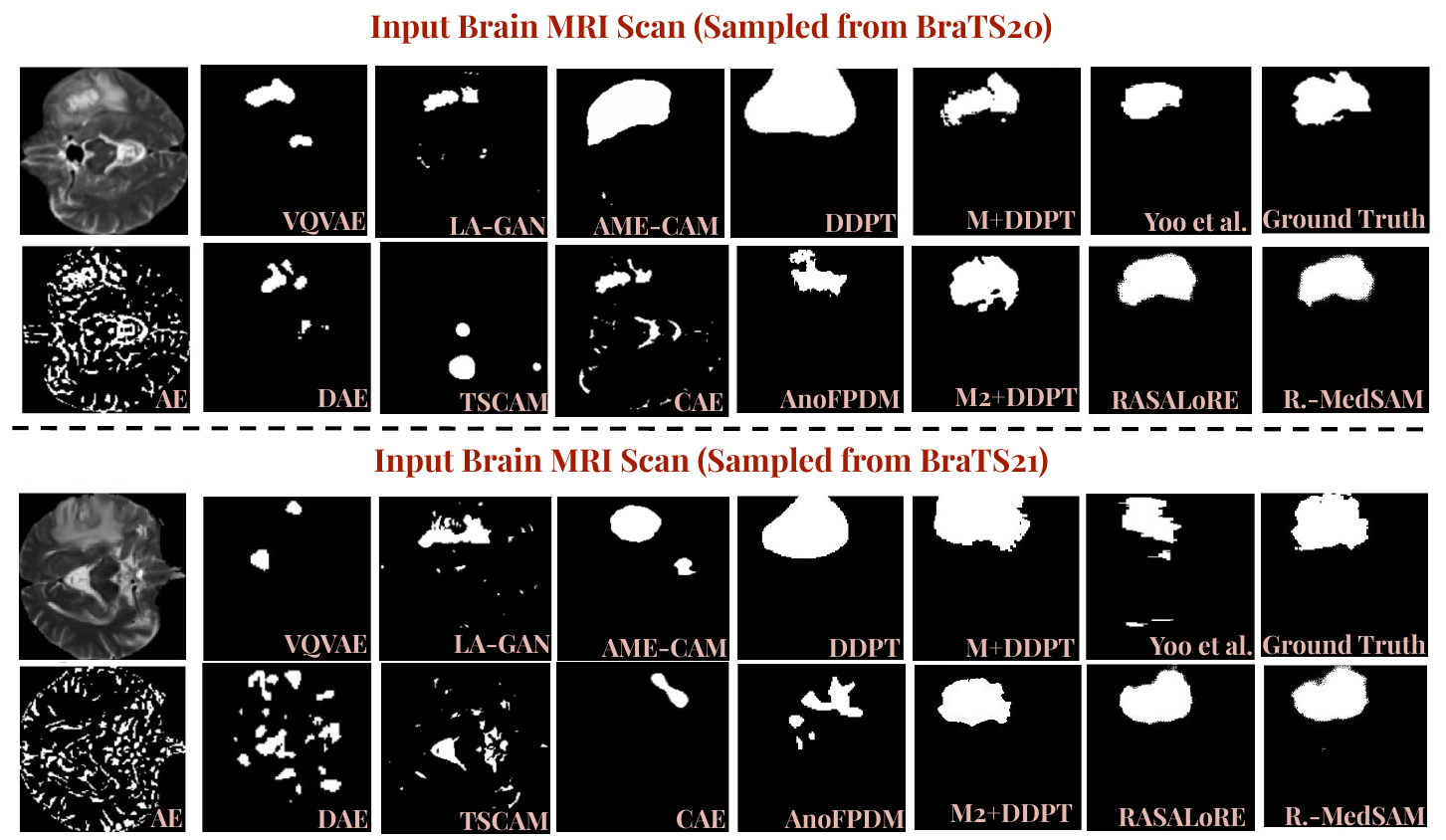}
			\vspace{0.2cm}
			\caption{Qualitative Comparison of Predicted Anomaly Mask from Different Methods. Abbreviations: M+DDPT = MedSAM+DDPT(box), M2+DDPT = MedSAM2+DDPT(box) and R.-MedSAM =  RASALoRE without MedSAM.}
		\label{fig:RASALoRE_Qualitative}
	\end{figure}	
	
	\textbf{Multimodality RASALoRE:}
	Table \ref{tab:multimodality_table} presents the quantitative performance of the proposed Multimodality RASALoRE. Here, the T2 modality has been used as a bridge modality. Results show that other modalities, which are usually not considered for anomaly detection due to low contrast and limited ability to capture fluid-containing structures (e.g., T1, T1ce), can still contribute meaningfully. In fact, using T1 and T1ce, our model achieves performance that is comparable and in some cases better than several comparative models (Table \ref{tab:Quantitative_same_domain_results}) operating on the T2 modality.
	
	\begin{table}[h]
		\centering
		\begin{tabular}{lcccccccc}
			\toprule
			& \multicolumn{2}{c}{T1} & \multicolumn{2}{c}{T2} & \multicolumn{2}{c}{Tice} & \multicolumn{2}{c}{Flair} \\
			\cmidrule(lr){2-3} \cmidrule(lr){4-5} \cmidrule(lr){6-7} \cmidrule(lr){8-9}
			Dataset & Dice & AUPRC & Dice & AUPRC & Dice & AUPRC & Dice & AUPRC \\
			\midrule
			BraTS20 & 65.13 & 66.90 & 71.82 & 77.08 & 66.77 & 68.86 & 72.42 & 75.62 \\
			BraTS21 & 63.24 & 62.87 & 68.57 & 73.55 & 67.65 & 67.67 & 69.53 & 74.23\\
			BraTS23 & 54.24 & 53.46 & 61.17 & 61.76 & 57.60 & 54.34 & 63.18 & 63.02 \\
			MSD & 56.19 & 60.66 & 67.31 & 73.63 & 61.18 & 64.04 & 69.20 & 74.59 \\
			\bottomrule
		\end{tabular}
		\vspace{0.5em}  
		\caption{Quantitative Results for Multi-Modality RASALoRE}
		\label{tab:multimodality_table}
	\end{table}
	
	\textbf{Ablation Studies and other experiments:} 
    Additional details and ablation are provided in Appendix~\ref{sec:rasalore} and Appendix~\ref{sec:ddpt_}.
	
	\vspace{-0.1in}
	\section{Conclusion}
	\label{sec:conclusion}
	
	We have proposed RASALoRE, a weakly supervised anomaly detection technique useful for anomaly segmentation in brain MRI scans, when ground-truth pixel-level annotations are unavailable. RASALoRE uses fixed candidate prompt point locations whose location-based random embeddings interact with suitable image-level intermediate feature representations, to provide sufficiently rich region-aware embeddings that elicit localized anomaly information from MRI scan images. We have also designed a weak mask generation technique, DDPT, which provides a weak supervisory signal for RASALoRE training. Our results showcase promising detection capabilities of RASALoRE on diverse BraTS-type datasets.

    \section{Acknowledgment}
    We thank the reviewers for their insightful comments which helped in improving our paper. We gratefully acknowledge Technocraft Centre of Applied Artificial Intelligence (TCAAI), IIT Bombay for their support through generous funding.
    
	\bibliography{egbib}
	
	\newpage
	\appendix
	\section{RASALoRE}
	\label{sec:rasalore}

	\subsection{Performance Across Different Prompt Embeddings}  
	\label{appendix:diff_prompt_emb}
	
	To evaluate the robustness of our approach against candidate prompt point (CPP) LoRE embeddings, we performed training and report performance across five different random CPP LoRE embeddings.
	The results are summarized in Table \ref{tab:dataset_performance}.  These results show comparable performance of our method across multiple random initializations of CPP LoRE embeddings. 
	\begin{table}[htbp]  
		\centering  
		\scriptsize  
		\renewcommand{\arraystretch}{0.9}  
		\begin{tabular}{cccccccc}  
			\hline  
			\textbf{Run} & \textbf{Dice \( \uparrow \)} & \textbf{AUROC \( \uparrow \) } & \textbf{AUPRC \( \uparrow \)} & \textbf{Precision \( \uparrow \) } & \textbf{Recall \( \uparrow \) } & \textbf{Specificity \( \uparrow \) } & \textbf{Jaccard \( \uparrow \) } \\  
			\hline  
			\multicolumn{8}{c}{\textbf{BraTS20}} \\  
			\hline  
			1 & 71.91\% & 94.65\% & 73.87\% & 71.18\% & 79.07\% & 96.97\% & 59.23\% \\  
			2 & 70.39\% & 92.87\% & 74.44\% & 73.57\% & 72.45\% & 97.70\% & 57.35\% \\  
			3 & 70.12\% & 93.26\% & 75.65\% & 77.18\% & 69.09\% & 98.10\% & 56.60\% \\  
			4 & 70.45\% & 93.63\% & 74.62\% & 72.84\% & 74.05\% & 97.26\% & 57.36\% \\  
			5 & 70.00\% & 93.32\% & 75.13\% & 75.73\% & 70.27\% & 97.97\% & 56.64\% \\  
			\textbf{Mean} & \textbf{70.57\%} & \textbf{93.55\%} & \textbf{74.74\%} & \textbf{74.10\%} & \textbf{72.99\%} & \textbf{97.60\%} & \textbf{57.44\%} \\  
			\textbf{STD}  & \(\boldsymbol{\pm}\) \textbf{0.77\%}  & \(\boldsymbol{\pm}\) \textbf{0.67\%}  & \(\boldsymbol{\pm}\) \textbf{0.68\%}  & \(\boldsymbol{\pm}\) \textbf{2.37\%}  & \(\boldsymbol{\pm}\) \textbf{3.90\%}  & \(\boldsymbol{\pm}\) \textbf{0.48\%} & \(\boldsymbol{\pm}\) \textbf{1.07\%} \\  
			\hline  
			
			\multicolumn{8}{c}{\textbf{BraTS21}} \\  
			\hline  
			1 & 71.60\% & 93.18\% & 73.88\% & 76.00\% & 72.25\% & 98.18\% & 58.02\% \\  
			2 & 70.69\% & 93.42\% & 75.58\% & 76.96\% & 69.76\% & 98.34\% & 56.87\% \\  
			3 & 70.96\% & 93.62\% & 75.84\% & 76.40\% & 70.74\% & 98.28\% & 57.26\% \\  
			4 & 70.61\% & 93.14\% & 75.26\% & 76.86\% & 69.65\% & 98.32\% & 56.75\% \\  
			5 & 70.39\% & 92.79\% & 74.71\% & 77.79\% & 68.34\% & 98.41\% & 56.43\% \\  
			\textbf{Mean} & \textbf{70.85\%} & \textbf{93.23\%} & \textbf{75.05\%} & \textbf{76.80\%} & \textbf{70.15\%} & \textbf{98.31\%} & \textbf{57.07\%} \\  
			\textbf{STD}  & \(\boldsymbol{\pm}\) \textbf{0.47\%}  & \(\boldsymbol{\pm}\) \textbf{0.31\%}  & \(\boldsymbol{\pm}\) \textbf{0.78\%}  & \(\boldsymbol{\pm}\) \textbf{0.67\%}  & \(\boldsymbol{\pm}\) \textbf{1.45\%}  & \(\boldsymbol{\pm}\) \textbf{0.08\%} & \(\boldsymbol{\pm}\) \textbf{0.61\%} \\  
			\hline  
			
			\multicolumn{8}{c}{\textbf{BraTS23}} \\  
			\hline  
			1 & 71.25\% & 94.54\% & 70.11\% & 67.70\% & 79.48\% & 97.00\% & 57.42\% \\  
			2 & 70.83\% & 94.68\% & 71.51\% & 69.40\% & 76.58\% & 97.32\% & 57.00\% \\  
			3 & 70.84\% & 94.86\% & 71.00\% & 66.98\% & 79.78\% & 96.80\% & 57.12\% \\  
			4 & 70.67\% & 94.41\% & 71.72\% & 69.37\% & 76.02\% & 97.38\% & 56.79\% \\  
			5 & 70.36\% & 94.76\% & 71.54\% & 68.34\% & 76.92\% & 97.24\% & 56.33\% \\  
			\textbf{Mean} & \textbf{70.79\%} & \textbf{94.65\%} & \textbf{71.18\%} & \textbf{68.36\%} & \textbf{77.76\%} & \textbf{97.15\%} & \textbf{56.93\%} \\  
			\textbf{STD}  & \(\boldsymbol{\pm}\) \textbf{0.32\%}  & \(\boldsymbol{\pm}\) \textbf{0.18\%}  & \(\boldsymbol{\pm}\) \textbf{0.65\%}  & \(\boldsymbol{\pm}\) \textbf{1.05\%}  & \(\boldsymbol{\pm}\) \textbf{1.74\%}  & \(\boldsymbol{\pm}\) \textbf{0.24\%} & \(\boldsymbol{\pm}\) \textbf{0.41\%} \\  
			\hline  		
			\multicolumn{8}{c}{\textbf{MSD}} \\  
			\hline  
			1 & 61.24\% & 89.24\% & 62.86\% & 66.68\% & 61.83\% & 97.23\% & 47.21\% \\  
			2 & 61.87\% & 89.82\% & 64.53\% & 65.69\% & 64.23\% & 96.72\% & 47.96\% \\  
			3 & 59.91\% & 88.51\% & 62.87\% & 66.75\% & 59.18\% & 97.41\% & 45.86\% \\  
			4 & 61.74\% & 89.45\% & 64.14\% & 65.80\% & 63.51\% & 96.89\% & 47.78\% \\  
			5 & 62.11\% & 92.83\% & 64.16\% & 67.32\% & 63.27\% & 97.08\% & 48.30\% \\  
			\textbf{Mean} & \textbf{61.37\%} & \textbf{89.97\%} & \textbf{63.71\%} & \textbf{66.45\%} & \textbf{62.40\%} & \textbf{97.07\%} & \textbf{47.42\%} \\  
			\textbf{STD}  & \(\boldsymbol{\pm}\) \textbf{0.88\%}  & \(\boldsymbol{\pm}\) \textbf{1.67\%}  & \(\boldsymbol{\pm}\) \textbf{0.79\%}  & \(\boldsymbol{\pm}\) \textbf{0.69\%}  & \(\boldsymbol{\pm}\) \textbf{2.00\%}  & \(\boldsymbol{\pm}\) \textbf{0.27\%} & \(\boldsymbol{\pm}\) \textbf{0.96\%} \\  
			\hline  
		\end{tabular}  
		\vspace{0.2cm}
		\caption{Performance Across Different Initializations of Candidate Prompt Embeddings}  
		\label{tab:dataset_performance}  
	\end{table}
	
	\subsection{Ablation based on Loss Components}
	\label{app:loss_ablation}
	\subsubsection{Ablation on hyperparameters $\alpha$ and $\beta$}
	\label{appenidx:loss_ablation}
	Table \ref{ablation} presents the results of an ablation study conducted on the BraTS20 dataset to evaluate the impact of third and fourth terms in $\mathcal{L}_{\text{Dec}}$ denoted by $\mathcal{L}^{3}_{\text{Dec}}$ and $\mathcal{L}^{4}_{\text{Dec}}$ respectively, which primarily focus on minimizing false positives. The study assesses different weights of the $\mathcal{L}^{3}_{\text{Dec}}$ and $\mathcal{L}^{4}_{\text{Dec}}$ terms. We first set $\beta=0$ and conducted experiments with different values of $\alpha$ for  $\mathcal{L}^{3}_{\text{Dec}}$ term. We find that $\alpha$ at 0.6 and 1.0 provide us better results.  Then we set $\alpha=0$ and analyze the impact of different choices of $\beta$ weight of $\mathcal{L}^{4}_{\text{Dec}}$ term. We observe that different values of weight parameter $\beta$ provide almost comparable results. Then, we conducted combination-based loss experiments by keeping $\alpha$ fixed at 0.6 and 0.1, while varying $\beta$. From these experiments, and from the corresponding qualitative results, we chose the combination $\alpha=0.6$ for $\mathcal{L}^{3}_{\text{Dec}}$ term and $\beta=0.6$ for $\mathcal{L}^{4}_{\text{Dec}}$ term. This choice  effectively captures both internal and boundary-specific features, also taking care of not impacting the true positive performance captured by the first two terms in  $\mathcal{L}_{\text{Dec}}$, while enhancing the true negative performance, thereby improving overall anomaly detection accuracy. We then used this choice for more experiments on all datasets (see Section \ref{Best_three_Decoder_Loss_Combination}), where we noted consistent observations. Hence $\alpha=0.6$ and $\beta=0.6$ were finally chosen for all our experiments.

	\begin{table}[h!]
		\centering
		\scriptsize
		\begin{tabular}{lccccccc}
			\toprule
			\textbf{Loss} & \textbf{Dice \( \uparrow \)} & \textbf{AUROC \( \uparrow \)} & \textbf{AUPRC \( \uparrow \)} & \textbf{Precision \( \uparrow \)} & \textbf{Recall \( \uparrow \)} & \textbf{Specificity \( \uparrow \)} & \textbf{Jaccard \( \uparrow \)} \\ 
			\midrule
			\multicolumn{8}{c}{\textbf{\(\alpha \times \mathcal{L}^{3}_{\text{Dec}}\)}} \\ 
			\midrule
			$\alpha$ = 0.2  & 70.34\% & 93.09\% & 73.52\% & 74.2\% & 72.84\% & 97.44\% & 57.17\% \\ 
			$\alpha$ = 0.4 & 70.51\% & 93.12\% & 73.72\% & 75.1\% & 72.01\% & 97.72\% & 57.27\% \\
			$\alpha$ = 0.6 & 70.62\% & 93.5\% & 73.94\% & 74.71\% & 72.6\% & 97.6\% & 57.5\% \\
			$\alpha$ = 0.8 & 69.92\% & 93.12\% & 73.96\% & 75.52\% & 70.61\% & 97.93\% & 56.64\% \\ 
			$\alpha$ = 1.0 & 70.62\% & 93.59\% & 73.96\% & 74.44\% & 72.84\% & 97.72\% & 57.56\% \\ 
			\midrule
			\multicolumn{8}{c}{\textbf{\(\beta \times \mathcal{L}^{4}_{\text{Dec}}\)}} \\ 
			\midrule
			$\beta$ = 0.2 & 70.16\% & 92.57\% & 72.26\% & 74.24\% & 72.05\% & 97.54\% & 56.93\% \\
			$\beta$ = 0.4 & 69.73\% & 92.09\% & 72.67\% & 75.20\% & 70.08\% & 98.18\% & 56.73\% \\
			$\beta$ = 0.6 & 69.47\% & 92.01\% & 71.76\% & 74.59\% & 70.25\% & 98.11\% & 56.58\% \\
			$\beta$ = 0.8 & 69.15\% & 92.04\% & 70.86\% & 74.43\% & 70.27\% & 97.50\% & 55.74\% \\		
			$\beta$ = 1.0 & 69.69\% & 91.42\% & 71.08\% & 74.65\% & 70.07\% & 98.15\% & 56.57\% \\ 
			\midrule
			\multicolumn{8}{c}{\textbf{Combination of \(\mathcal{L}^{3}_{\text{Dec}}\) and \(\mathcal{L}^{4}_{\text{Dec}}\)}} \\ 
			\midrule
			$\alpha = 0.6, \beta = 0.2$ & 69.62\% & 92.96\% & 75.96\% & \textbf{78.46\%} & 66.89\% & \textbf{98.52\%} & 56.00\% \\
			$\alpha = 0.6, \beta = 0.4$ & 70.71\% & 93.72\% & 76.13\% & 75.25\% & 71.98\% & 97.72\% & 57.53\% \\
			$\alpha = 0.6, \beta = 0.6$ & \textbf{71.91\%} & \textbf{94.65\%} & 73.87\% & 71.18\% & \textbf{79.07\%} & 96.97\% & \textbf{59.23\%} \\
			$\alpha = 0.6, \beta = 0.8$ & 69.36\% & 93.33\% & 75.10\% & 75.10\% & 69.97\% & 97.80\% & 55.90\% \\
			$\alpha = 0.6, \beta = 1.0$ & 69.36\% & 93.33\% & 75.10\% & 75.10\% & 69.97\% & 97.80\% & 55.90\% \\
			$\alpha = 1.0, \beta = 0.2$ & 69.95\% & 93.08\% & 75.64\% & 75.90\% & 70.00\% & 97.83\% & 56.63\% \\
			$\alpha = 1.0, \beta = 0.4$ & 70.88\% & 93.65\% & \textbf{76.44\%} & 76.27\% & 71.09\% & 98.05\% & 57.60\% \\
			$\alpha = 1.0, \beta = 0.6$ & 69.60\% & 92.29\% & 74.77\% & 75.29\% & 69.06\% & 97.99\% & 56.39\% \\
			$\alpha = 1.0, \beta = 0.8$ & 69.06\% & 92.69\% & 73.96\% & 75.19\% & 68.87\% & 97.61\% & 55.51\% \\
			$\alpha = 1.0, \beta = 1.0$ & 69.15\% & 92.11\% & 74.14\% & 76.67\% & 67.11\% & \textbf{98.14\%} & 55.74\% \\

			\bottomrule
		\end{tabular}
		\vspace{0.2cm}
		\caption{Performance comparison based on different combinations of $\alpha$ and $\beta$ in the $\mathcal{L}_{\text{Dec}}$ loss on the BraTS20 dataset.}
		\label{ablation}
	\end{table}

	\subsubsection{Ablation based on Best Three Decoder Loss Combination Across All Datasets}
	\label{Best_three_Decoder_Loss_Combination}
	
	\begin{table}[h!]
		\centering
		\scriptsize
		\begin{tabular}{cccccccc}
			\hline
			\multicolumn{8}{c}{\textbf{$0.6 \times \mathcal{L}^{3}_{\text{Dec}}$ + $ 0.6 \times \mathcal{L}^{4}_{\text{Dec}}$}} \\ \hline
			\textbf{Tested on} & \textbf{Dice \( \uparrow \)} & \textbf{AUROC \( \uparrow \)} & \textbf{AUPRC \( \uparrow \)} & \textbf{Precision \( \uparrow \)} & \textbf{Recall \( \uparrow \)} & \textbf{Specificity \( \uparrow \)} & \textbf{Jaccard \( \uparrow \)} \\ \hline
			BraTS20          & 71.91\% & 94.65\% & 73.87\% & 71.18\% & 79.07\% & 96.97\% & 59.23\% \\ 
			BraTS21          & 71.60\% & 93.18\% & 73.88\% & 76.00\% & 72.25\% & 98.18\% & 58.02\% \\ 
			BraTS23          & 71.25\% & 94.54\% & 70.11\% & 67.70\% & 79.48\% & 97.00\% & 57.42\% \\ 
			MSD              & 61.24\% & 89.24\% & 62.86\% & 66.68\% & 61.83\% & 97.23\% & 47.21\% \\ \hline
			\multicolumn{8}{c}{\textbf{$\mathcal{L}^{3}_{\text{Dec}}$}} \\ \hline
			\textbf{Tested on} & \textbf{Dice \( \uparrow \)} & \textbf{AUROC \( \uparrow \)} & \textbf{AUPRC \( \uparrow \)} & \textbf{Precision \( \uparrow \)} & \textbf{Recall \( \uparrow \)} & \textbf{Specificity \( \uparrow \)} & \textbf{Jaccard \( \uparrow \)} \\ \hline
			BraTS20          & 70.62\% & 93.59\% & 73.96\% & 74.44\% & 72.84\% & 97.72\% & 57.56\% \\ 
			BraTS21          & 71.60\% & 92.94\% & 74.52\% & 77.01\% & 71.03\% & 98.34\% & 57.96\% \\ 
			BraTS23          & 71.21\% & 94.89\% & 70.77\% & 68.61\% & 78.39\% & 97.19\% & 57.42\% \\ 
			MSD              & 61.41\% & 89.17\% & 62.76\% & 67.13\% & 61.77\% & 97.22\% & 47.35\% \\ \hline
			\multicolumn{8}{c}{$\mathcal{L}^{4}_{\text{Dec}}$} \\ \hline
			\textbf{Tested on} & \textbf{Dice \( \uparrow \)} & \textbf{AUROC \( \uparrow \)} & \textbf{AUPRC \( \uparrow \)} & \textbf{Precision \( \uparrow \)} & \textbf{Recall \( \uparrow \)} & \textbf{Specificity \( \uparrow \)} & \textbf{Jaccard \( \uparrow \)} \\ \hline
			BraTS20          & 69.69\% & 91.42\% & 71.08\% & 74.65\% & 70.07\% & 98.15\% & 56.57\% \\ 
			BraTS21          & 70.96\% & 92.62\% & 73.75\% & 76.78\% & 70.35\% & 98.22\% & 57.17\% \\ 
			BraTS23          & 71.11\% & 95.12\% & 70.79\% & 67.47\% & 79.66\% & 96.91\% & 57.26\% \\ 
			MSD              & 62.40\% & 89.07\% & 63.15\% & 67.52\% & 63.35\% & 97.23\% & 48.30\% \\ \hline
		\end{tabular}
		\vspace{0.2cm}
		\caption{Performance obtained by best three combinations of $\alpha$, $\beta$ of $\mathcal{L}_{\text{Dec}}$ loss across various datasets.}
		\label{tab:best_three_loss}
	\end{table}
	
	Based on the ablation study in Table \ref{ablation}, we selected the three top-performing loss combinations, namely $(\alpha=1, \beta=0)$, $(\alpha=0, \beta=1)$ and $(\alpha=0.6, \beta=0.6)$ for further evaluation. To assess their consistency across different datasets, we conducted experiments on all four datasets. Using Table \ref{tab:best_three_loss} and qualitative performance, we observed that the combination of $\mathcal{L}^{3}_{\text{Dec}}$ and $\mathcal{L}^{4}_{\text{Dec}}$ loss components of $\mathcal{L}_{\text{Dec}}$, with $\alpha$ and $\beta$ both set to 0.6, consistently performed better than the other two combinations. Therefore, we chose this loss configuration for our final experiments. 
	
	\subsection{Qualitative Results on BraTS23 and MSD Datasets}
	
	Figure~\ref{fig:RASALoRE_Qualitative} presents the qualitative results of our proposed RASALoRE method on the BraTS23 and MSD datasets. As evident from the figure, RASALoRE demonstrates superior segmentation performance, accurately delineating the anomalies compared to other competing methods.                                                                                                               
	\begin{figure}[htbp]
		\centering
		\includegraphics[width=\linewidth]{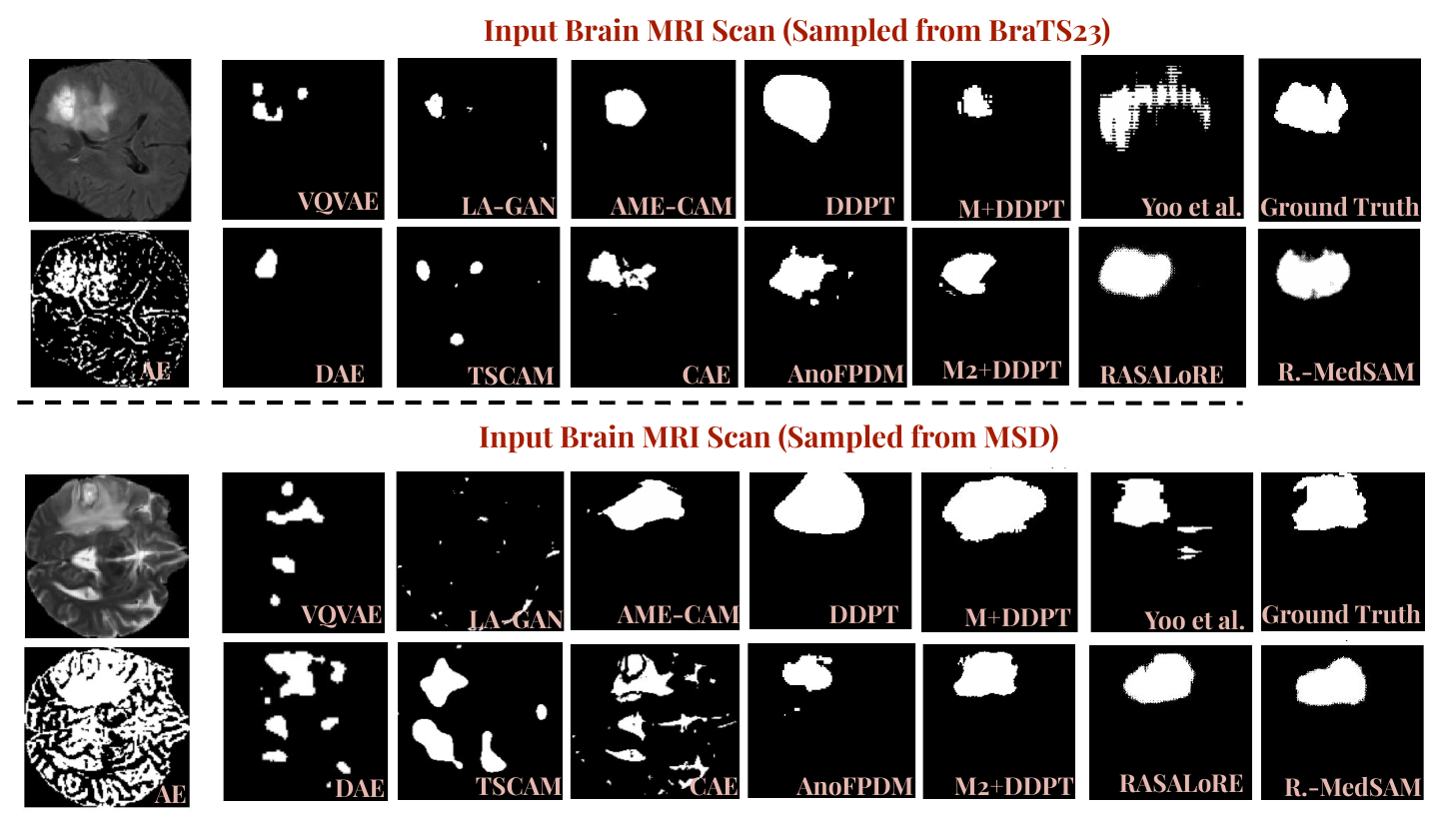}
		\vspace{-0.3cm}
		\caption{Qualitative Comparison of Predicted Anomaly Mask from Different Methods. Abbreviation: M+DDPT = MedSAM+DDPT(box), M2+DDPT = MedSAM2+DDPT(box) and R.-MedSAM =  RASALoRE without MedSAM}
		\label{fig:RASALoRE_Qualitative}
	\end{figure}	
	
	\subsection{MedSAM and MedSAM2 Inference}
	\label{MedSAM_inference}
	MedSAM and MedSAM2 are user-prompt driven methods. Since we are operating in an unsupervised setup, we have exclusively utilized DDPT-based weak masks for prompt guidance. MedSAM and MedSAM2 support both bounding box and point-based prompts. We conducted inference with both settings and tabulated the results in Table 1 in the paper. For bounding box selection, we identified the rows and columns where the DDPT-based weak mask was nonzero. By determining the left-most and right-most column numbers, as well as the top-most and bottom-most row numbers, we defined the coordinates of the bounding box. This bounding box was then used as the input prompt for MedSAM's and MedSAM2's box-based inference. For point-based inference, we first extracted the bounding box from the DDPT mask and randomly selected 10 points within it, which served as prompts during inference. Since bounding box prompts yielded better performance for MedSAM, we adopted that in the design of our loss function (discussed in Section 2 in the paper). 

	\subsection{Qualitative Results of Multi-modality RASALoRE}
	
\begin{figure}[htbp]
		\centering
		\includegraphics[width=\linewidth]{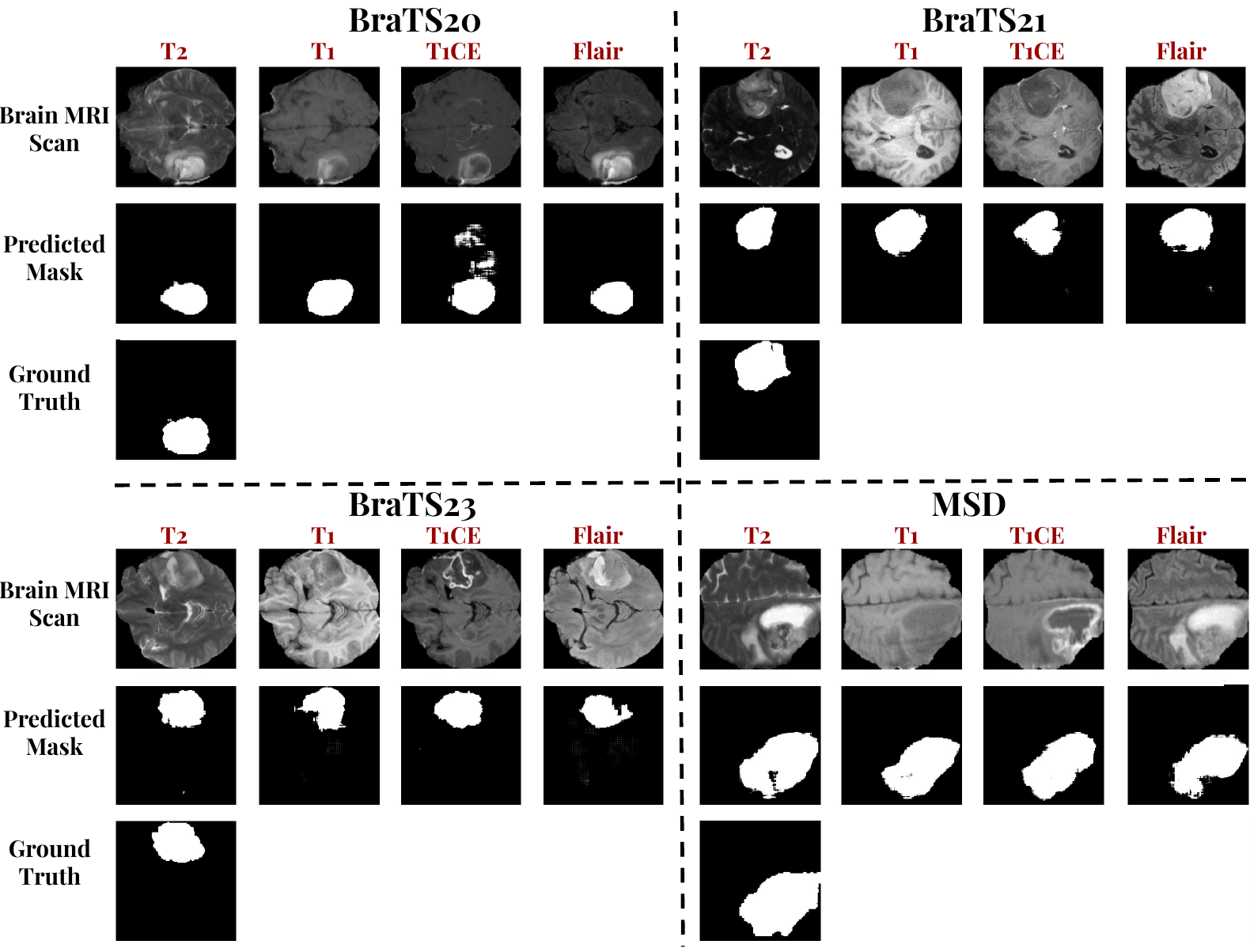}
		\vspace{-0.3cm}
		\caption{Qualitative Results of Multi-Modality RASALoRE}
		\label{fig:MM_RASALoRE_Qualitative}
	\end{figure}

	Figure~\ref{fig:MM_RASALoRE_Qualitative} presents the qualitative results of our multi-modality RASALoRE across various datasets. It can be observed that the predicted masks obtained using the T1 or T1ce modality are comparable to those derived from the T2 and FLAIR modalities. This highlights the effectiveness of our proposed model and the underlying idea of using location-based random embeddings, which enables the model to achieve consistent performance across modalities with minimal additional parameters.

	\subsection{Comparison with Diffusion Based Models}
	\label{app:diffusion}
	
	Table \ref{comparative_segmentation_results} (reproduced in Table \ref{comparative_segmentation_results_again}) and Table \ref{comparative_segmentation_results_21} show comparison of DDPT with various models like AE, DDPM, pDDPM, cDDPM, and mDDPM on BraTS20 and BraTS21 datasets, which highlights the significant performance advantage of our DDPT-based approach compared to diffusion models. Also these diffusion based baseline models require substantial computational costs and prolonged training times, which make them less suitable for resource-constrained scenarios. Since our DDPT-based method consistently outperforms these diffusion models in terms of both segmentation accuracy and computational efficiency on BraTS20 and BraTS21 datasets, we have not conducted further experiments for comparing these diffusion based models directly with RASALoRE. Instead, our comparative analysis focuses on evaluating RASALoRE against DDPT, given that DDPT already establishes itself as a better alternative to diffusion-based methods.
	
	\begin{table}[!h]
		\small
		\centering
		\begin{tabular}{cccccc}
			\hline
			& & \multicolumn{2}{c}{\textbf{Test on: BraTS21}} & \multicolumn{2}{c}{\textbf{Test on: MSLUB}} \\
			\cline{3-6}
			\textbf{Method} & \textbf{Train on} & \textbf{Dice} & \textbf{AUPRC} & \textbf{Dice} & \textbf{AUPRC}\\
			\hline
			AE  & BraTS20 & 7.08$\pm$1.57 & 14.26$\pm$3.19 & 2.72$\pm$0.45 & 1.82$\pm$0.13 \\
			DDPM& BraTS20 & 28.91$\pm$2.74 & 48.21$\pm$3.32 & 1.22$\pm$0.32 & 2.52$\pm$0.07\\
			
			pDDPM & BraTS20 & 24.03$\pm$0.81 & 44.42$\pm$1.50 &  0.94$\pm$0.12 & 2.23$\pm$0.11\\
			cDDPM  & BraTS20 & 27.59$\pm$2.43 & 39.86$\pm$4.09 & 1.35$\pm$0.18 & 1.86$\pm$0.24\\
			mDDPM & BraTS20 & 25.18$\pm$2.27 & 46.37$\pm$4.24 & 0.97$\pm$0.17 & 2.34$\pm$0.16 \\
			\hline
			DDPT(threshold)& BraTS20 & \textbf{52.44$\pm$0.69} & \textbf{60.69$\pm$0.73} & \textbf{5.67$\pm$0.09} & \textbf{10.43$\pm$0.39}\\ 
			DDPT(Class Guided)& BraTS20 & {\color{blue}{57.74$\pm$0.56}} & {\color{blue}{64.71$\pm$0.59}} &  {\color{blue}8.58$\pm$0.42} &  {\color{blue} 12.23$\pm$0.58}\\ 
			\hline
		\end{tabular}
		\vspace{0.2cm}
		\caption{Comparative Analysis of Segmentation Results. The results in this table are fetched from Table \ref{comparative_segmentation_results}.}
		\label{comparative_segmentation_results_again}
	\end{table}
	
	\subsection{Remarks on ATLAS and MSLUB Datasets}
	\label{appendix:atlas_mslub}
	
	From our experiments with the DDPT model on the MSLUB dataset \cite{lesjak2018novel} (see Table~\ref{comparative_segmentation_results_again}), we observed that the low scores obtained are inconclusive and should not be considered indicative of meaningful results. As discussed in the experimental section \ref{Analysis_MSLUB}, our analysis of the MSLUB dataset shows that such Dice scores can be achieved using noise alone. The same observation applies to the ATLAS dataset \cite{rohlfing2010sri24}. Therefore, we have excluded these datasets from further comparison. It is likely that these datasets require more careful consideration and a deeper investigation into their data characteristics to yield meaningful insights.
	
	\begin{figure}[h!]
		\centering
		\begin{overpic}[width=0.8\textwidth]{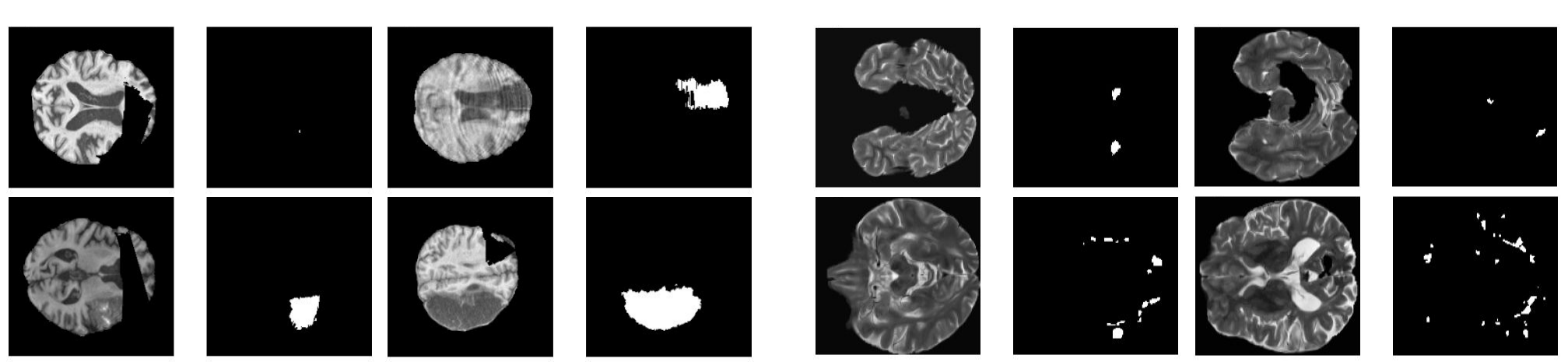}
			\put(1, 22){\scalebox{0.4}{\textcolor{red}{\textbf{Brain MRI Scan}}}}
			\put(16, 22){\scalebox{0.4}{\textcolor{red}{\textbf{GT Mask}}}}
			\put(25, 22){\scalebox{0.4}{\textcolor{red}{\textbf{Brain MRI Scan}}}}
			\put(40, 22){\scalebox{0.4}{\textcolor{red}{\textbf{GT Mask}}}}
			\put(53, 22){\scalebox{0.4}{\textcolor{red}{\textbf{Brain MRI Scan}}}}
			\put(67, 22){\scalebox{0.4}{\textcolor{red}{\textbf{GT Mask}}}}
			\put(77, 22){\scalebox{0.4}{\textcolor{red}{\textbf{Brain MRI Scan}}}}
			\put(91, 22){\scalebox{0.4}{\textcolor{red}{\textbf{GT Mask}}}}
			\put(20, -3){\scalebox{1}{\textcolor{black}{\textbf{ATLAS}}}}
			\put(70, -3){\scalebox{1}{\textcolor{black}{\textbf{MSLUB}}}}
		\end{overpic}
		\vspace{0.5cm}
		\caption{Samples from ATLAS and MSLUB Datasets indicating low quality of the data.}
		\label{fig:ATLAS_MSLUB}
	\end{figure}
	
	In Figure \ref{fig:ATLAS_MSLUB}, we provide sample images from both datasets, illustrating their complexity, corruption, and quality issues, which hinder the ability of learning algorithms to perform effectively on these datasets.
	
	\subsection{MedSAM vs SAM Mask Comparison}
	\label{appendix:MedSAMvsSAM}
\begin{figure}[!h]
		\centering
		\includegraphics[width=\textwidth]{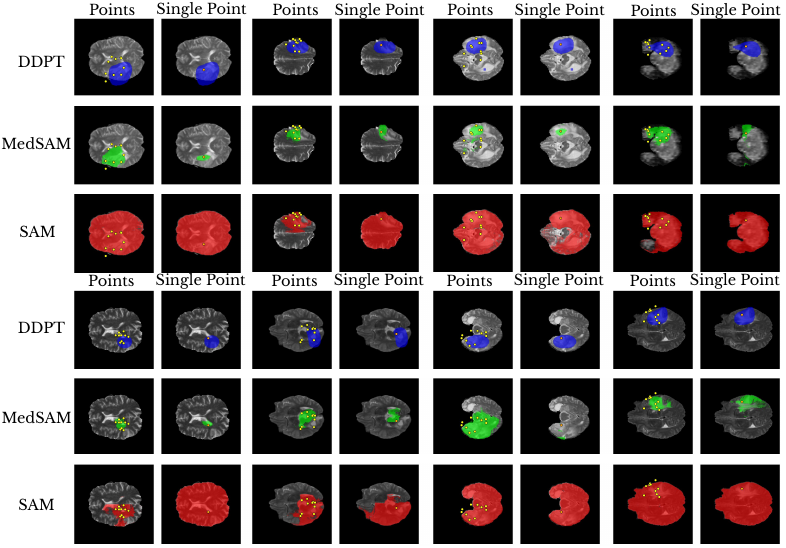}
		\vspace{0.5cm}
		\caption{Qualitative comparison of segmentation results when SAM and MedSAM are prompted using points from the weak mask obtained by DDPT. These results highlight the limitations of the original SAM in accurately localizing anomalies compared to MedSAM when using weak prompts.}
		\label{fig:Compare_MedSAM_SAM}
	\end{figure}

	As shown in Figure~\ref{fig:Compare_MedSAM_SAM}, MedSAM demonstrates better alignment with the anomaly regions (in green color) when prompted with DDPT’s weak masks (in blue color). In contrast, the original SAM often fails to precisely localize the pathology, producing fragmented or misaligned segmentations (in red color). This highlights MedSAM’s improved adaptability to medical imaging contexts. These observations highlight the critical role of domain-specific pretraining in medical image segmentation and motivate our adoption of MedSAM-generated weak masks in the design of the proposed loss function.
	
	\subsection{Architecture Details}
	\label{appendix:RASALoRE_arch}
	
	In this section, we provide architectural details for each component of RASALoRE, including information on the layers used, as well as the input and output shapes of each module.
	
	Recall that candidate prompt points are obtained using a grid of shape $\sqrt{k} \times \sqrt{k}$. When this grid is overlaid on the input image of shape $h \times w$, we obtain appropriate $k$ candidate locations associated with image, which are then associated with their respective LoRE. 
	We now provide the details of other components of the RASALoRE architecture.

	\textbf{Image Encoder}
	
	\begin{scriptsize} 
		\begin{longtable}{@{}lll@{}}
			\toprule
			\textbf{}       & \textbf{Input Shape: (Batch, 1, h, w)}                                                                          & \textbf{}      \\ \midrule
			\textbf{Layer}       & \textbf{Details}                                                                          & \textbf{Output Shape}      \\ \midrule
			Encoder Block 1      & Conv2d(1 $\to$ 32, kernel=3x3, stride=1, padding=1)                                        & (Batch, 32, h, w)         \\
			& Conv2d(32 $\to$ 32, kernel=3x3, stride=1, padding=1)                                       & (Batch, 32, h, w)         \\
			& ReLU                                                                                      & (Batch, 32, h, w)         \\
			& LayerNorm                                                                              & (Batch, 32, h, w)     \\
			& Dropout2d(p=0.1)                                                                          & (Batch, 32, h, w)     \\
			& MaxPool2d(kernel=2x2, stride=2)                                                           & (Batch, 32, h/2, w/2)     \\ \midrule
			Encoder Block 2      & Conv2d(32 $\to$ 64, kernel=3x3, stride=1, padding=1)                                       & (Batch, 64, h, w)     \\
			& Conv2d(32 $\to$ 32, kernel=3x3, stride=1, padding=1)                                       & (Batch, 64, h, w)         \\
			& ReLU                                                                                      & (Batch, 64, h, w)         \\
			& LayerNorm                                                                              & (Batch, 64, h, w)     \\
			& Dropout2d(p=0.1)                                                                          & (Batch, 64, h, w)     \\
			& MaxPool2d(kernel=2x2, stride=2)                                                           & (Batch, 64, h/2, w/2)     \\ \midrule
			Encoder Block 3      & Conv2d(64 $\to$ 128, kernel=3x3, stride=1, padding=1)                                      & (Batch, 128, h, w)    \\
			& Conv2d(128 $\to$ 128, kernel=3x3, stride=1, padding=1)                                     & (Batch, 128, h, w)    \\
			& ReLU                                                                                      & (Batch, 128, h, w)    \\
			& LayerNorm                                                                              & (Batch, 128, h, w)     \\
			& Dropout2d(p=0.1)                                                                          & (Batch, 128, h, w)     \\
			& MaxPool2d(kernel=2x2, stride=2)                                                           & (Batch, 128, h/2, w/2)     \\ \midrule
			Encoder Block 4      & Conv2d(128 $\to$ 256, kernel=3x3, stride=1, padding=1)                                     & (Batch, 256, h/2, w/2)    \\
			& Conv2d(256 $\to$ 256, kernel=3x3, stride=1, padding=1)                                     & (Batch, 256, h/2, w/2)    \\
			& ReLU                                                                                      & (Batch, 256, h/2, w/2)    \\
			& LayerNorm                                                                              & (Batch, 256, h/2, w/2)     \\
			& Dropout2d(p=0.1)                                                                          & (Batch, 256, h/2, w/2)     \\
			& MaxPool2d(kernel=2x2, stride=2)                                                           & (Batch, 256, h/4, w/4)     \\ \bottomrule
		\end{longtable}
	\end{scriptsize} 

	\textbf{Refiner}
	\begin{scriptsize} 
		\begin{longtable}{@{}lll@{}}
			\toprule
			\textbf{}          & \textbf{Input Shape: (Batch, 32, h/2, w/2)}                                                                       & 
			\textbf{}      \\ \midrule
			\textbf{Layer}          & \textbf{Details}                                                                       & \textbf{Output Shape}      \\ \midrule
			Encoder Block 1      & Conv2d(1 $\to$ 32, kernel=3x3, stride=1, padding=1)                                        & (Batch, 32, h, w)         \\
			& Conv2d(32 $\to$ 32, kernel=3x3, stride=1, padding=1)                                       & (Batch, 32, h, w)         \\
			& ReLU                                                                                      & (Batch, 32, h, w)         \\
			& MaxPool2d(kernel=2x2, stride=2)                                                           & (Batch, 32, h/2, w/2)     \\
			& LayerNorm                                                                              & (Batch, 32, h/2, w/2)     \\
			& Dropout2d(p=0.1)                                                                          & (Batch, 32, h/2, w/2)     \\ \midrule
			Conv                   & Conv2d(32 $\to$ 64, kernel=3x3, stride=1, padding=1)                                    & (Batch, 64, h/4, w/4)    \\
			BatchNorm + ReLU       & BatchNorm2d(64) + ReLU                                                                  & (Batch, 64, h/4, w/4)    \\ \midrule
			UpConv 1               & ConvTranspose2d(64 $\to$ 128, kernel=3x3, stride=1, padding=1)                          & (Batch, 128, h/2, w/2)   \\
			& BatchNorm2d(128) + ReLU                                                                 & (Batch, 128, h/2, w/2)   \\ \midrule
			UpConv 2               & ConvTranspose2d(128 $\to$ 256, kernel=3x3, stride=1, padding=1)                         & (Batch, 256, h, w)   \\
			& BatchNorm2d(256) + ReLU                                                                 & (Batch, 256, h, w)   \\ \midrule
			Skip Conv              & Conv2d(32 $\to$ 256, kernel=1x1, stride=1)                                              & (Batch, 256, h, w)   \\
			Global Avg Pool        & AvgPool2d(kernel=4, stride=2)                                                           & (Batch, 256, h/4, w/4)        \\ \bottomrule
		\end{longtable}
	\end{scriptsize} 
	
	\textbf{Region Aware Spatial Attention}
	\begin{scriptsize} 
		\begin{longtable}{@{}lll@{}}
			\toprule
			\textbf{}           & 
			\textbf{Input Image Refined Features Shape: (Batch, 256, h/4, w/4)}                                                                      & 
			\textbf{} \\ \midrule
			\textbf{}           & 
			\textbf{Input Candidate Embedding Shape (Batch, k, d)}                                                                      & 
			\textbf{} \\ \midrule
			\textbf{Layer}           & \textbf{Details}                                                                      & \textbf{Output Shape} \\ \midrule
			Cross-Attention          & MultiheadAttention(256 $\to$ 256) 4 Heads                                                     & (Batch, k, 256)          \\ \midrule                                                      &                       \\ \bottomrule
		\end{longtable}
	\end{scriptsize} 
	
	\textbf{Feed Forward Network}
	\begin{scriptsize} 
		\begin{longtable}{@{}lll@{}}
			\toprule
			\textbf{}           & 
			\textbf{Input Spatial Point Embedding Shape: (Batch, k, d)}                                                                      & 
			\textbf{} \\ \midrule
			\textbf{Layer}           & \textbf{Details}                                                                      & \textbf{Output Shape} \\ \midrule
			Feedforward Block        & Linear(256 $\to$ 128) → ReLU → LayerNorm(128) → Linear(128 $\to$ 64) → ReLU            & (Batch, k, 1)       \\
			& LayerNorm(64) → Sigmoid                                                               &                       \\ \bottomrule
		\end{longtable}
	\end{scriptsize} 
	
	\textbf{Decoder}
	\begin{scriptsize} 
		\begin{longtable}{@{}lll@{}}
			\toprule
			\textbf{}           & 
			\textbf{Input Image Refined Features Shape: (Batch, 256, h/4, w/4)}                                                                      & 
			\textbf{} \\ \midrule
			\textbf{}           & 
			\textbf{Input Spatial Point Embedding Shape (Batch, k, d)}                                                                      & 
			\textbf{} \\ \midrule
			\textbf{Layer}           & \textbf{Details}                                                                      & \textbf{Output Shape} \\ \midrule
			Cross-Attention          & MultiheadAttention(256 $\to$ 256) 4 Heads                                                     & (Batch, 4096, 256)          \\ \midrule
			Norm 1                  & LayerNorm(256)                                                                         & (Batch, 4096, 256)          \\ \midrule
			Upsample 1               & ConvTranspose2d(1024 $\to$ 128, kernel=4x4, stride=2, padding=1) → ReLU                & (Batch, 128, h/4, w/4) \\
			Upsample 2               & ConvTranspose2d(128 $\to$ 64, kernel=4x4, stride=2, padding=1) → ReLU                  & (Batch, 64, h/2, w/2) \\
			Final Conv               & Conv2d(64 $\to$ 1, kernel=1x1, stride=1)                                              & (Batch, 1, h, w)      \\ \bottomrule
		\end{longtable}
	\end{scriptsize} 
	
	\newpage
	\section{Discriminative Dual Prompt Tuning}
	\label{sec:ddpt_}
	\subsection{DDPT Recap}
	
\begin{figure}[h]
		\centering
		\includegraphics[width=0.9\linewidth]{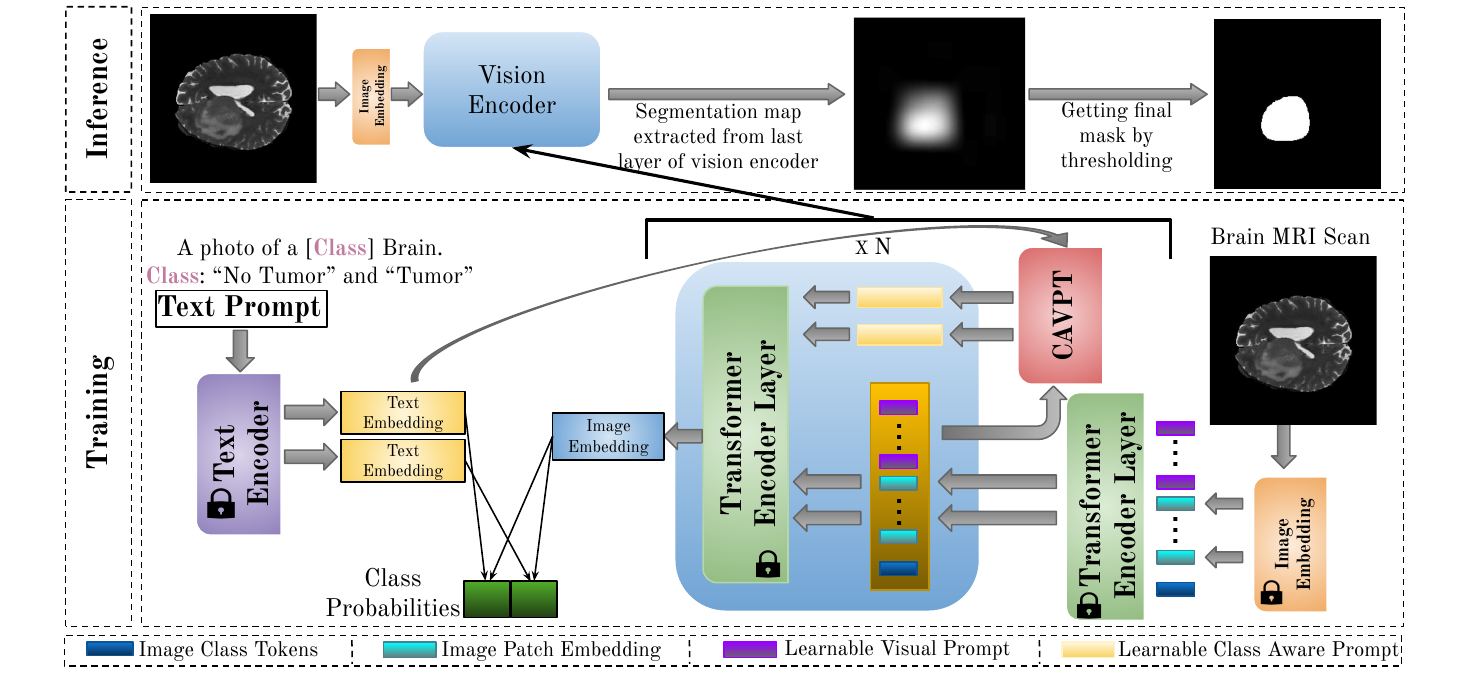}
		\vspace{0.1cm}
		\caption{Overview of DDPT Architecture}

	\label{fig:Architecture}
	\end{figure}

	As discussed in Section 2.1 of the paper, we consider a discriminative vision language model (CLIP) \cite{radford2021learning} by fine-tuning both the text and visual prompts while keeping the image and text encoders frozen. Our objective is to extract attention maps from the last layer of the visual transformer within the image encoder that potentially localizes regions of interest.  The extracted attention maps are then reshaped to align with the spatial dimensions of the original input image, offering a localized representation of potential anomalies (Figure \ref{fig:Architecture}).
	
	\subsection{Experiments and Results using DDPT}
	\subsubsection{Experimental Setup and Dataset Preprocessing}
	We conducted our experiments using PyTorch 2.0.1 framework in a Linux environment, with an NVIDIA GeForce RTX 4090 graphics card with 24GB of memory. We employed the SGD optimizer with a learning rate of 0.01, weight decay set at $5\times 10^{-4}$, and momentum at $0.9$ for model training as used in \cite{xing2023dual}. The temperature coefficient $\tau$ is initialized as 0.07 from \cite{wu2018unsupervised}.
	
	For our experiments, we utilized T2 modality from BraTS20 \cite{bakas2017advancing,bakas2018identifying,menze2014multimodal}, BraTS21 \cite{baid2021rsna,bakas2017advancing,menze2014multimodal}, MSLUB \cite{lesjak2018novel} datasets. These datasets offer a diverse range of brain MRI scans for analysis  in NIfTI format. BraTS20 dataset consists of 369 volumes, BraTS21 contains 1251 volumes and MSLUB comprises 30 volumes. The following preprocessing steps were applied to all datasets. Initially, we removed the initial 15 frames and the last 15 frames of each volume. This preprocessing step was undertaken to focus the analysis on regions containing at least minimal relevant information. It is noted that a majority of the initial and final frames do not contain brain tissues. This would ensure that the segmentation process focuses on the most important areas. Subsequently, the NIfTI volume files from the dataset are converted into $256 \times 256$ sized images. Since we are aware of image level information, we split the image set into two parts, one for healthy images and one for unhealthy. Furthermore, we divided each dataset into 80\% for training and 10\% for validation and 10\% for testing.
	In our experiments, we explored the impact of different data shot sizes on model adaptation. We denote one shot as a single (healthy image, unhealthy image) pair. Specifically, we tested shot sizes of 64, 256, 1024, and 10,000 for each dataset aiming to enhance the model's performance. Our findings indicate that shot sizes of 10,000 yielded the most favorable results, allowing us to achieve improved performance.
	
	\subsubsection{Training and Inference}
	The model is first trained on the training split for a classification task, say to predict if the input image is healthy or not. After this classification training, during inference, each image along with a prompt is fed into the trained model. Attention maps are then extracted from the last layer of the vision transformer encoder. These attention maps serve as the basis for predicting the location of anomalies in the brain. The attention map is normalized so that values are in the range [0,1] and thresholded (with threshold $0.5$) to result in the corresponding anomaly segment mask. Metrics are then computed based on these masks to assess the model's performance.
	
	Two training methods are employed in the experiments based on: 
	\begin{itemize}\setlength\itemsep{0em}
	\item \textbf{H1:} Handcrafted prompts are utilized for the text encoder, with prompts structured as ``a photo of a \{ \} brain.", where the class information `Tumor' and `No Tumor' are used in the place of \{ \}.
	\item \textbf{L1:} A prompt learner, inspired by CoOP \cite{zhou2022learning}, is utilized, where information of both classes `Tumor' and `No Tumor' are included during training.
	\end{itemize}
	
	Two inference methods are employed specifically for anomalous images: 
	\begin{itemize}\setlength\itemsep{0em}
	\item \textbf{H2:} The same handcrafted prompts used during training are employed, with only the `Tumor' class specified during inference.
	\item \textbf{L2:} Learned prompts from the prompt learner during training are employed, with the `Tumor' class specified during inference. 
	\end{itemize}
	
	Thus for each type of training (with fixed handcrafted prompts \textbf{H1} and with learnable prompts \textbf{L1}), we have conducted inference with the two inference methods \textbf{H2} and \textbf{L2} resulting in four different combinations. Note that when training is done with \textbf{H1} and inference is done with \textbf{L2}, only a randomly initialized prompt vector is used in inference. During inference, each image from the test dataset is processed alongside the prompt. Attention maps from the last layer of the vision encoder are extracted during forward propagation and are used to evaluate segmentation task metrics (see Figure \ref{fig:Architecture}).
	
	\subsubsection{Comparison based on the number of shots used for training}
	In this experiment, models are trained over different number of shots for 60 epochs on the datasets. Recall that 1 shot refers to a pair of positive and negative images. The 10000 shot experiment is also done over 100 epochs, since it achieved better performance with 60 epochs.  
	For the simplicity of comparison, we use only H1-L2 type of training and inference. We use the same dataset for both training and testing in this experiment. We report results in Table \ref{Table1_shots} on BraTS20 and on BraTS21 datasets. From Table \ref{Table1_shots}, we notice that as we increase the number of shots of training data, the model gets better at producing the segmentation mask. However the increase in the performance saturates for higher number of shots. Using these experiments, we fixed the number of shots to be 10000 in our experiments. From Table \ref{Table1_shots}, we also note that the classification accuracy of unseen test images when using only 10000 shots is around 91\% for BraTS20 dataset and $\approx$92\% for BraTS21 dataset.

	\begin{table}[h]
	\centering
	\begin{tabular}{ccccccc}
		\hline
		Dataset  & Shots   & Epochs  & Class Acc.  & Dice Score & IOU & AUPRC \\
		\hline
		\hline
		BraTS20 & 64 & 60 & 75.90\% & 0.1317302 & 0.0421849 & 0.2522678 \\
		& 256 & 60 & 82.90\% & 0.4152587 & 0.1377230 & 0.4983816 \\
		& 1024 & 60 & 87.80\% & 0.5571311 & 0.2963292 & 0.6158307\\
		& 10000 & 60 & 91.90\% & 0.6034027 & 0.3220559 & 0.6658436 \\
		& 10000 & 100 & 92.40\% & 0.6055508 & 0.3266677 & 0.6662489 \\
		
		\hline
		\hline
		
		BraTS21 & 64 & 60 & 79.60\%	& 0.2116216	& 0.0916265	& 0.3973538\\
		& 256 & 60 & 86.70\% & 0.5147872 & 0.1785913 & 0.5805774 \\
		& 1024 & 60 & 88.30\% &	0.5725719 &	0.3028589 &	0.6234630\\
		& 10000 & 60 & 91.20\% & 0.6119194 & 0.3333350 & 0.6718447\\
		& 10000 & 100 & 91.20\% & 0.6139209 & 0.3241785 & 0.6742342\\
		\hline
	\end{tabular}
	\vspace{0.3cm}
	\caption{Results for segmentation metrics for models trained on different number of shots for both the BraTS20 and BraTS21 datasets.}
	\label{Table1_shots}
	\end{table}

	\subsubsection{Comparison based on the Training and Inference Methods}
	For this experiment, we take the best performing model (10000 shots, 100 epochs), and we compare the metrics for the different combinations of training and inference methods. We present the results in Table \ref{Table_train_inf}.
	
	\begin{table}[htbp]
	\centering
	
	\begin{tabular}{ccccccc}
		\hline
		Dataset  & Method & Shots   & Epochs  &  Dice Score & IOU & AUPRC \\
		\hline
		\hline
		BraTS20 & H1-H2	& 10000 & 100 & 0.6055508 & 0.3266677 & 0.6662489 \\
		& L1-H2	& 10000 &	100 & 0.5943685 & 0.3095980 & 0.6555493\\
		& L1-L2	& 10000 &	100 & 0.5943466 & 0.3098367 & 0.6554740\\

		\hline
		\hline
		
		BraTS21 & H1-H2 & 10000 & 100 & 0.613920 & 0.3241785 & 0.6742342 \\
		& L1-H2	& 10000 &	100 & 0.5963832 & 0.3126123 & 0.6555436\\
		& L1-L2	& 10000 &	100 & 0.5963929 & 0.3126953 & 0.6555216\\
		
		\hline
	\end{tabular}
	\vspace{0.3cm}
	\caption{Results for segmentation metrics for different combinations of training and inference methods, on both BraTS20 and BraTS21 datasets.}
	\label{Table_train_inf}
	\end{table}

	A common observation from Table \ref{Table_train_inf} over both the datasets is that the training method involving the hand crafted prompts (H1) (using the prompt ``a photo of a \{ \} brain”), is performing consistently more better than incorporating the CoOp learnable vectors (L1). This can be explained by the fact that CLIP is pre-trained over a vast amount of natural text-image pairs, and hence its exposure to medical terms might be quite low, which would limit its capability when using the learnable prompts. However hand crafting them makes it easier in such domains. 
	
	\subsubsection{Cross Domain Segmentation Results}
	In Table \ref{comparative_segmentation_results}, we present a comprehensive comparison of various segmentation methods evaluated on different datasets, focusing on two evaluation metrics namely, the Dice coefficient and area under precision-recall curve (AUPRC). 
	
	\textbf{Comparative methods:} Recent methods including DDPM \cite{ho2020denoising}, pDDPM \cite{behrendt2024patched}, cDDPM \cite{behrendt2023guided}, and mDDPM \cite{iqbal2023unsupervised} were initially trained using the IXI dataset (after extensive pre-processing), as reported in Behrendt et al. \cite{behrendt2024patched}. Their aim was generation-based segmentation, where they utilized the IXI dataset containing only healthy brain images for training and BraTS21 and MSLUB which contains both type of images (healthy and unhealthy) for testing.
	
	To provide a comparative analysis with our approach, we trained AE \cite{luo2023unsupervised}, DDPM \cite{ho2020denoising}, pDDPM \cite{behrendt2024patched}, cDDPM \cite{behrendt2023guided}, and mDDPM \cite{iqbal2023unsupervised} models on the healthy brain images obtained from the BraTS20 dataset. The experimental settings are same as those utilized in the corresponding  papers \cite{behrendt2024patched,behrendt2023guided,iqbal2023unsupervised,luo2023unsupervised,ho2020denoising}. The BraTS20 dataset comprises 369 volumes, which were partitioned into five folds, with 300 volumes allocated for training and 69 for validation. For evaluation purposes, we employed both the BraTS21 and MSLUB datasets, which consist of 1251 and 30 volumes, respectively. Each trained fold was subsequently evaluated on the unhealthy images obtained from BraTS21 and MSLUB datasets to construct a corresponding healthy image. The segmentation mask was obtained by normalizing and thresholding (at 0.5) the difference between the generated healthy image and the input unhealthy image. Note that during training of these methods, image-level label information was considered to be known. Then five different trained models were used for evaluation to compute mean and standard deviations of the Dice coefficient and AUPRC metrics over these five models as in \cite{behrendt2024patched}. Note that, AE \cite{luo2023unsupervised} is for 3D volumetric segmentation, so we converted it to the 2D framework for comparative analysis.
	
	\textbf{Our method:} Recall that based on the type of prompts used for training (\textbf{H1} and \textbf{L1}), we use two different models for training. In each model, we utilized 10,000 shots and 100 epochs of training. 
	
	Our model undergoes training using the BraTS20 dataset, which contains both healthy and unhealthy images. Before training, from the NIfTI volume files in the dataset, frames are extracted and converted into $256 \times 256$ sized images without extensive pre-processing. Then this set of extracted images is split into healthy and unhealthy images based on known image-level label information. Our models are trained for classification tasks and evaluated on unhealthy images obtained from both the BraTS21 and MSLUB datasets (similar to comparative methods). Specifically, for the BraTS21 and MSLUB datasets, we utilize only the unhealthy images (containing abnormalities) extracted from 1251 NIfTI volumes and 30 NIfTI volumes, respectively. During evaluation, a test image is fed along with a text prompt (handcrafted or random) into the trained model, and the last layer activations from the final transformer's encoder are used to generate a pixel-wise segmentation information grid for the test image. This grid is further resized to the original image shape, normalized to be in the range $[0,1]$ and then a threshold of $0.5$ is applied on the grid to create predicted segmentation labels. We evaluate the performance using dice coefficient and AUPRC(Area Under the Precision-Recall Curve) metrics.
	
	Additionally we have also explored a classifier guided segmentation approach. Recall from Table \ref{Table1_shots}, that our model can provide a classification label for the test image with the classification accuracy of around $90\%$. From our experiments, we observed that the classification decision from our model is based on the relative proportion of anomalous regions present in the unhealthy test images. Thus based on the classification decision, we have ignored those images which get classified as healthy image by our model, and we have computed dice coefficient and AUPRC only on those test images which have been classified as unhealthy images by our model. Though this might lead to removal of a few unhealthy images from our final test set, it indicates that the amount of anomalous region pixels present in the discarded test image is not sufficient to guide the classifier's decision, and hence cannot be found to be reliable information for the downstream segmentation task as well.    
	
	\textbf{Comparative analysis:} Table \ref{comparative_segmentation_results} shows the results of our segmentation approach using fixed handcrafted prompts (\textbf{H1}), trained on the BraTS20 dataset. We opted to compare models trained with the (\textbf{H1}) prompt since it consistently yielded the most favorable results across the various experiments conducted.
	Subsequently, we evaluated our model's performance on the BraTS21 and MSLUB datasets. Notably from the results in Table \ref{comparative_segmentation_results}, our approach outperforms others on BraTS21 dataset, achieving the highest mean dice coefficients of 52.44 and AUPRC score of 60.69. 
	Additionally, the model evaluated with classifier guidance provides a mean dice coefficient of 57.74 and AUPRC score of 64.71. Note that this improvement was possible because of the model's ability to discard images with prediction masks having insignificant anomalous regions. We further observed that using classifier guidance, unhealthy test images wrongly predicted as healthy, typically have anomalous pixel counts of less than 200.
	
	\begin{table}[h]
	\small
	\centering
	\begin{tabular}{cccccc}
		\hline
		& & \multicolumn{2}{c}{\textbf{BraTS21}} & \multicolumn{2}{c}{\textbf{MSLUB}} \\
		\cline{3-6}
		\textbf{Method} & \textbf{Train on} & \textbf{Dice} & \textbf{AUPRC} & \textbf{Dice} & \textbf{AUPRC}\\
		\hline
		AE  & BraTS20 & 7.08$\pm$1.57 & 14.26$\pm$3.19 & 2.72$\pm$0.45 & 1.82$\pm$0.13 \\
		DDPM& BraTS20 & 28.91$\pm$2.74 & 48.21$\pm$3.32 & 1.22$\pm$0.32 & 2.52$\pm$0.07\\
		pDDPM & BraTS20 & 24.03$\pm$0.81 & 44.42$\pm$1.50 &  0.94$\pm$0.12 & 2.23$\pm$0.11\\
		cDDPM  & BraTS20 & 27.59$\pm$2.43 & 39.86$\pm$4.09 & 1.35$\pm$0.18 & 1.86$\pm$0.24\\
		mDDPM & BraTS20 & 25.18$\pm$2.27 & 46.37$\pm$4.24 & 0.97$\pm$0.17 & 2.34$\pm$0.16 \\
		\hline
		Ours(threshold)& BraTS20 & \textbf{52.44$\pm$0.69} & \textbf{60.69$\pm$0.73} & \textbf{5.67$\pm$0.09} & \textbf{10.43$\pm$0.39}\\ 
		Ours(Class Guided)& BraTS20 & {\color{blue}{57.74$\pm$0.56}} & {\color{blue}{64.71$\pm$0.59}} &  {\color{blue}8.58$\pm$0.42} &  {\color{blue} 12.23$\pm$0.58}\\ 
		\hline
	\end{tabular}
	\vspace{0.2cm}
	\caption{Comparative Analysis of Segmentation Results. The table presents segmentation performance metrics for various methods trained on BraTS20 dataset and evaluated on both BraTS21 and MSLUB datasets. For our method with class guidance (highlighted in blue), certain test images were excluded from dice coefficient and AUPRC computations based on classification decisions. Results are presented as mean $\pm$ standard deviation.}
	\label{comparative_segmentation_results}
	\end{table}

	\begin{table}[!h]
	\small
	\centering
	\begin{tabular}{cccccc}
		\hline
		& & \multicolumn{2}{c}{\textbf{BraTS20}} & \multicolumn{2}{c}{\textbf{MSLUB}} \\
		\cline{3-6}
		\textbf{Method} & \textbf{Train on} & \textbf{Dice} & \textbf{AUPRC} & \textbf{Dice} & \textbf{AUPRC}\\
		\hline
		AE \cite{luo2023unsupervised} & BraTS21 & 15.10  & 22.86 & 3.21 & 1.90 \\
		DDPM \cite{ho2020denoising}& BraTS21 & 26.19 & 45.51 & 1.25 & 2.30 \\
		pDDPM \cite{behrendt2024patched} & BraTS21 & 24.67 & 47.05 & 1.01 & 2.13\\
		cDDPM \cite{behrendt2023guided} & BraTS21 & 37.66 & 48.77 & 1.89 & 1.74 \\
		mDDPM \cite{iqbal2023unsupervised} & BraTS21 & 26.27 & 51.14 & 1.09 & 2.07\\
		\hline
		Ours(threshold) & BraTS21 & \textbf{54.92} & \textbf{62.56} & \textbf{5.35} & \textbf{9.63} \\ 
		Ours(Class Guided)& BraTS21 & {\color{blue}{62.72}} & {\color{blue}{68.21}} &  {\color{blue}9.43} &  {\color{blue}12.07}\\ 
		\hline
	\end{tabular}
	\vspace{0.2cm}
	\caption{Comparative Analysis of Segmentation Results. The table presents segmentation performance metrics for various methods trained on BraTS21 dataset and evaluated on both BraTS20 and MSLUB datasets. For our method with class guidance (highlighted in blue), certain test images were excluded from dice coefficient and AUPRC computations based on classification decisions. Results are presented as mean.}
	\label{comparative_segmentation_results_21}
	\end{table}

	It is important to note that for both our proposed method and comparative methods such as DDPM, pDDPM, cDDPM, mDDPM, and AE, a threshold of 0.5 was consistently applied.
	
	Table \ref{comparative_segmentation_results_21} presents a comparative analysis of results obtained when the model is trained using the BraTS21 dataset and evaluated on both the BraTS20 and MSLUB datasets. The BraTS21 dataset consists of 1251 volumes, with 1151 volumes allocated for training and 100 for validation after partitioning. It is noteworthy that the results in Table \ref{comparative_segmentation_results_21} were obtained by training the model on BraTS21 with only one fold, as training for 5 folds would be prohibitively time-consuming due to the dataset size. Additionally, our model outperforms all comparative methods by a considerable margin, achieving a mean dice coefficient of 54.92\% and an AUPRC score of 62.56\%. For the MSLUB dataset, we observed low values of performance metrics, similar to the previous case.

	\textbf{Analysis of MSLUB Dataset:} 
	\label{Analysis_MSLUB}
	We conducted a separate analysis on the MSLUB data, calculating the dice score between the original masks and randomly generated binary masks. The resulting dice score was approximately 4\%. Considering that our model in Table \ref{comparative_segmentation_results} achieved dice scores around 5\%, it is important to note that these slight improvements may be attributed to random noise rather than genuine model efficacy. This suggests that the methods in Table \ref{comparative_segmentation_results} may not be entirely suitable for handling the complexity of the MSLUB dataset.

	\textbf{Training Time Comparison:} Our method demonstrates significantly shorter training times compared to comparative models such as DDPM, pDDPM, cDDPM, mDDPM, and AE, as illustrated in Table \ref{training_time}. This indicates the efficiency of our approach in terms of computational resources and time investment.
	\begin{table}[!h]
	\centering
	\begin{tabular}{ccc}
		\hline
		\textbf{Training Dataset} & \multicolumn{1}{c}{\textbf{BraTS20}} & \multicolumn{1}{c}{\textbf{BraTS21}} \\
		\hline
		\textbf{Model} & \textbf{Training Time} & \textbf{Training Time} \\
		\hline
		AE \cite{luo2023unsupervised}& 6 & 9 \\
		DDPM  \cite{ho2020denoising} & 10 & 18 \\
		pDDPM \cite{behrendt2024patched}& 10 & 18 \\
		cDDPM \cite{behrendt2023guided}& 12 & 21 \\
		mDDPM \cite{iqbal2023unsupervised}& 12 & 21 \\
		\hline
		Ours & \textbf{3} & \textbf{5}\\
		\hline
	\end{tabular}
	\vspace{0.2cm}
	\caption{Training time taken for each fold by Different Models (In hours)}
	\label{training_time}
	\end{table}

\begin{figure}[!h]
	\centering
	\includegraphics[width=1.0\textwidth]{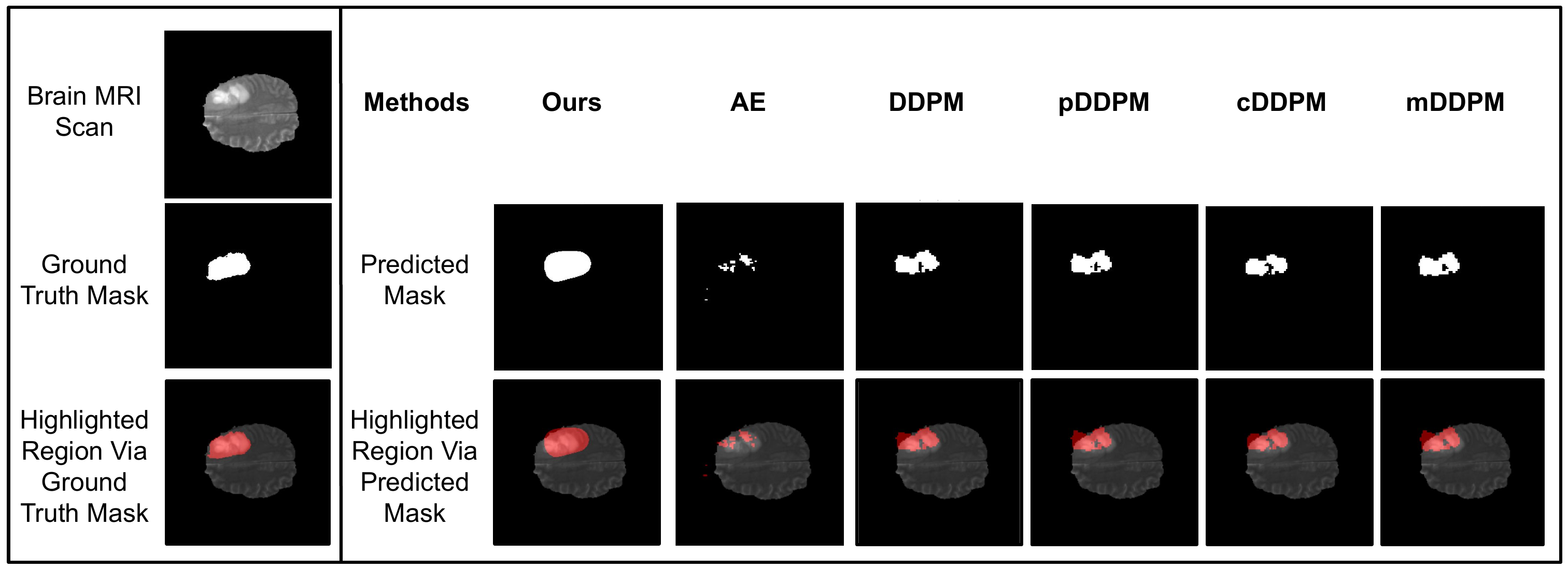}
	\vspace{0.2cm}
	\caption{Qualitative Results for BraTS21 Dataset}
	\label{fig:BraTS21_qual}
	\end{figure}

	\textbf{Qualitative results:} The efficiency of our segmentation approach is demonstrated through qualitative results depicted in Figure \ref{fig:BraTS21_qual}. These results, obtained from the model trained on the BraTS20 dataset and tested on the BraTS21 dataset, showcase the effectiveness of our method in accurately capturing anomalous regions. Additionally, qualitative results from comparative methods are also included in Figure \ref{fig:BraTS21_qual}.
	
	\textbf{Ablation Studies:}
	
	\textbf{(a) Training and Inference with Different Types of Prompts:} Table \ref{tab:10000_BraTS20_21_different_training_type} illustrates the segmentation outcomes on the BraTS21 dataset, employing various combinations of training and inference prompts. It is essential to emphasize that these findings stem from a single-fold training regimen, where the model is trained on BraTS20 and evaluated on BraTS21. Notably, the metrics for all training types exhibit similarities. However, we observed that the \textbf{H1} and \textbf{L2} combination yields slightly better overall performance across different experiments. Hence, we have opted to utilize that combination.

	\textbf{(b) Determining thresholding limit:} For this experiment, the models were trained with 256 samples over 60 epochs, to evaluate which threshold for the attention maps gave the best segmentation maps. We show the results using different combinations of the training and inference prompts. From the results in Table \ref{threshold_exp}, we consistently saw that the threshold of $0.5$ performed the best, which has been used for all the experiments in this work. The evaluation has been done over both BraTS20 and BraTS21 datasets. 
	
	\begin{table}[h]
	\centering
	\begin{tabular}{ccccccc}
		\hline
		\multicolumn{1}{c}{\textbf{Training}} & \multicolumn{1}{c}{\textbf{Inference}} & \multicolumn{1}{c}{\textbf{Dice Score}} & \multicolumn{1}{c}{\textbf{AUPRC}}\\
		\multicolumn{1}{c}{\textbf{Type}} & \multicolumn{1}{c}{\textbf{Prompt}} & \multicolumn{1}{c}{\textbf{}} & \multicolumn{1}{c}{\textbf{}}  \\
		\hline
		H1 & H2 & 51.80 & 59.99 \\ \hline
		H1 & L2 & 51.82 & 60.03 \\ \hline
		L1 & H2 & 51.66 & 60.41 \\ \hline
		L1 & L2 & 51.66 & 60.40 \\ \hline
	\end{tabular}
	\vspace{0.2cm}
	\caption{Segmentation Results with Different Types of Prompts}
	\label{tab:10000_BraTS20_21_different_training_type}
	\end{table}
	
	\begin{table}[h]
	\centering
	\begin{tabular}{ccccccc}
		\hline
		\multicolumn{1}{c}{\textbf{Dataset}} & \multicolumn{1}{c}{\textbf{Prompt combination}} & \multicolumn{3}{c}{\textbf{Threshold}} \\
		\cline{3-5}
		\multicolumn{1}{c}{\textbf{}} & \multicolumn{1}{c}{\textbf{}} & \multicolumn{1}{c}{\textbf{0.4}} & \multicolumn{1}{c}{\textbf{0.5}}& \multicolumn{1}{c}{\textbf{0.6}} \\
		
		\hline
		BraTs20 & H1-L1 & 39.88 & 41.52 & 39.86 \\ \hline
		BraTs20 & H1-L2 & 39.87 & 41.54 & 39.88 \\ \hline

		BraTs21 & H1-L1 & 50.88 & 51.47 & 48.67 \\ \hline
		BraTs21 & H1-L2 & 50.88 & 51.48 & 48.67 \\ \hline
		
	\end{tabular}
	\vspace{0.2cm}
	\caption{Segmentation DICE scores for different thresholding values over the attention maps (256 samples, 60 epochs, 200 pixel count threshold)}
	\label{threshold_exp}
	\end{table}
	
	\textbf{(c) Cross Domain Generalization with Anomalous Pixel Count:} For this experiment, the models were trained with 10,000 samples and \textbf{H1} text prompt, for 100 epochs. we considered an additional threshold on the number of pixels labeled as anomalous, based on the grid of pixels obtained after thresholding the activations (with threshold $0.5$) from the final encoder of the vision transformer.  Table \ref{Cross_domian_with_thresh} presents a concise summary of segmentation results in cross-domain adaptation with different thresholds based on anomalous pixel count. This leads to consider only those test images with a higher degree of anomalous pixel presence. The results show competitive results on both BraTS20 and BraTS21 datasets.

	\textbf{(d) Analysis of Weight Parameter $\eta$:} Table \ref{alpha} presents a comparative analysis of segmentation results obtained by training with a single fold for different values of the weight parameter, $\eta$. As $\eta$ increases from 0.1 to 0.3, both the Dice coefficient and AUPRC metrics demonstrate improvements. However, when $\eta$ increases further from 0.3 to 0.7, both metrics show a slight decline. Given that the best performance is achieved at $\eta = 0.3$, we selected this value for all experiments.
	
	\begin{table}[h]
	\centering
	\begin{tabular}{ccccc}
		\hline
		\textbf{Pixel count Threshold} & \textbf{Train On} & \textbf{Test On} & \textbf{Dice Coefficient} & \textbf{AUPRC} \\
		\hline
		0 & BraTS20 & BraTS21 & 52.44 $\pm$ 0.69 & 60.69 $\pm$ 0.73\\ 
		200 & BraTS20 & BraTS21 & 59.34 $\pm$ 0.78 & 65.35 $\pm$ 0.78\\ 
		400 & BraTS20 & BraTS21 & 63.30 $\pm$ 0.83 & 67.79 $\pm$ 0.79\\
		\hline
		\hline
		0 & BraTS21 & BraTS20 & 54.92 $\pm$ 0.58 & 62.56 $\pm$ 0.45 \\ 
		200 & BraTS21 & BraTS20 & 61.20 $\pm$ 0.66  & 66.90 $\pm$ 0.53 \\ 
		400 & BraTS21 & BraTS20 & 64.64 $\pm$ 0.71 & 69.05 $\pm$ 0.58 \\ 
		\hline
	\end{tabular}
	\vspace{0.2cm}
	\caption{Comparative Analysis of Segmentation Results in Cross-Domain Adaptation with Anomalous Pixel Count (Threshold use for predication map is 0.5)}
	\label{Cross_domian_with_thresh}
	\end{table}

	\begin{table}[h]
	\centering
	\begin{tabular}{ccccc}
		\hline
		\textbf{Eta($\eta$)} & \textbf{Train On} & \textbf{Test On} & \textbf{Dice Coefficient} & \textbf{AUPRC} \\
		\hline
		0.1 & BraTS20 & BraTS21 & 51.65 & 60.36 \\
		0.3 & BraTS20 & BraTS21 & 51.80 & 60.69 \\ 
		0.5 & BraTS20 & BraTS21 & 51.66 & 59.92 \\
		0.7 & BraTS20 & BraTS21 & 50.89 & 60.23 \\
		\hline
	\end{tabular}
	\vspace{0.2cm}
	\caption{Comparative Analysis of Segmentation Results on different $\eta$ values}
	\label{alpha}
	\end{table}

	\subsection{Summary of DDPT}
	We introduce a discriminative vision-text prompt tuning approach, which leverages a pre-trained CLIP model and image-level labels for anomaly detection in brain MRI images. 
	By emphasizing the importance of our discriminative approach over traditional generative methods, we offer significant promise for real-time clinical applications and resource-constrained environments, further advancing medical image analysis.

\end{document}